\documentclass[10pt,twocolumn,letterpaper]{article}

\usepackage{sec_CVPR2025/cvpr}              %

\definecolor{cvprblue}{rgb}{0.21,0.49,0.74}
\usepackage[pagebackref,breaklinks,colorlinks,allcolors=cvprblue]{hyperref}

\def\paperID{18593} %
\def\confName{CVPR}
\def\confYear{2025}

\title{A Closer Look at Time Steps is Worthy of Triple Speed-Up for Diffusion Model Training}

\usepackage{bbding}
\usepackage{colortbl}
\usepackage{amsthm}
\usepackage{wrapfig}
\usepackage{multirow}
\usepackage{adjustbox}
\usepackage{xcolor}
\definecolor{RoyalBlue}{RGB}{65,105,225}
\definecolor{RedOrange}{RGB}{255,69,0}
\definecolor{AccelerationColor}{RGB}{146,208,80}
\definecolor{DecelerationColor}{RGB}{0,176,240}
\definecolor{ConvergenceColor}{RGB}{237,125,49}
\definecolor{Weighting}{RGB}{184,201,233}
\definecolor{Sampling}{RGB}{255,166,166}
\usepackage{dsfont}
\hypersetup{
    colorlinks=true,
    linkcolor=red,
    citecolor=cyan,
    filecolor=magenta,      
    urlcolor=magenta,
    }
\newtheorem{theorem}{Theorem}
\newtheorem{lemma}{Lemma}
\newtheorem{remark}{Remark}
\usepackage{listings}

\newcommand{\methodname}{SpeeD }
\newcommand{\deltaname}{process increment }
\newcommand{\Deltaname}{Process increment }

\newlength\savewidth\newcommand\shline{\noalign{\global\savewidth\arrayrulewidth
  \global\arrayrulewidth 1pt}\hline\noalign{\global\arrayrulewidth\savewidth}}

\definecolor{defaultcolor}{HTML}{E8E2F7}
\newcommand{\default}[1]{\cellcolor{defaultcolor}{#1}}

\definecolor{baselinecolor}{gray}{.9}
\newcommand{\baseline}[1]{\cellcolor{baselinecolor}{#1}}

\newcommand{\tablestyle}[2]{\setlength{\tabcolsep}{#1}\renewcommand{\arraystretch}{#2}\centering\footnotesize}
\renewcommand{\paragraph}[1]{\vspace{1.25mm}\noindent\textbf{#1}}

\author{
  Kai Wang$^{1*}$,\; Mingjia Shi$^{1*}$,\; Yukun Zhou$^{1,2}$,\;Zekai Li$^{1}$,\; Zhihang Yuan$^{3}$,\; Yuzhang Shang$^{4}$,\; \\ \ Xiaojiang Peng$^{2\dagger}$\textbf{,}\; Hanwang Zhang$^{5}$\textbf{,}\;Yang You$^{1}$
   \\
  $^{1}$National University of Singapore \quad
  $^{2}$Shenzhen Technology University \quad
  $^{3}$Infinigence-AI \quad \\
  $^{4}$Illinois Institute of Technology \quad
  $^{5}$Nanyang Technological University\\
\ Code: \href{https://github.com/NUS-HPC-AI-Lab/SpeeD}{NUS-HPC-AI-Lab/SpeeD}
}

\begin{document}
\maketitle
\renewcommand{\thefootnote}{}
\footnotetext[0]{$^{*}$equal contribution, $^\dagger$corresponding author.}
\begin{abstract}

Training diffusion models is always a computation-intensive task. In this paper, we introduce a novel speed-up method for diffusion model training, called \textit{SpeeD}, which is based on a closer look at time steps.
Our key findings are:
i) Time steps can be empirically divided into acceleration, deceleration, and convergence areas based on the process increment.
ii) These time steps are imbalanced, with many concentrated in the convergence area.
iii) The concentrated steps provide limited benefits for diffusion training.
To address this, we design an asymmetric sampling strategy that reduces the frequency of steps from the convergence area while increasing the sampling probability in other areas. Additionally, we propose a weighting strategy to emphasize the importance of time steps with rapid-change process increments.
As a plug-and-play and architecture-agnostic approach, \textit{SpeeD} consistently achieves \textbf{3$\times$} acceleration across various diffusion architectures, datasets, and tasks. Notably, due to its simple design, our approach significantly reduces the cost of diffusion model training with minimal overhead.
Our research enables more researchers to train diffusion models at a lower cost.

\end{abstract}
\section{Introduction}

Training diffusion models is not usually affordable for many researchers, especially for ones in academia. For example, DALL$\cdot$E 2~\citep{Dalle-2} needs 40K A100 GPU days and Sora~\citep{Sora} at least necessitates 126K H100 GPU days. Therefore, accelerating the training for diffusion models has become urgent for broader generative AI and related applications.

\begin{figure}
    \centering
    \includegraphics[width=0.5\textwidth]{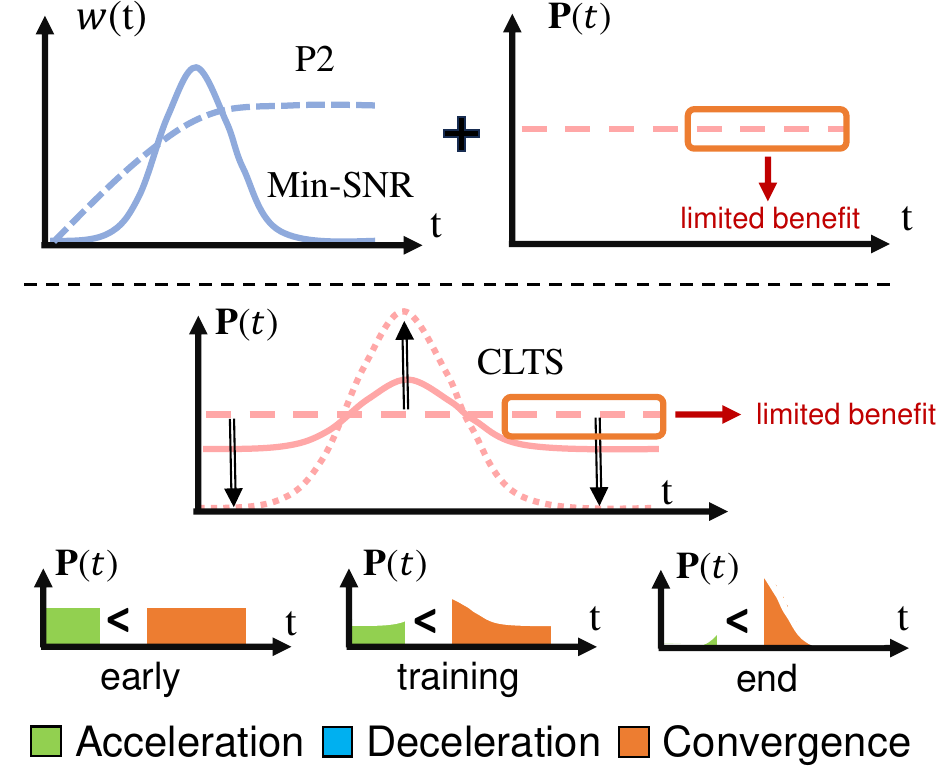}
    \caption{Re-weighting and re-sampling methods can't eliminate the redundancy and under-sample issues. $w(t)$ and $\mathbf{P}(t)$ are respectively the weighting and sampling curve. The probability of {convergence} area being sampled remains, while the one of {acceleration} is reduced faster.}
    \vspace{-10pt}
    \label{fig:1b}
\end{figure}

Recently, some acceleration methods for diffusion training focus on time steps, primarily using re-weighting and re-sampling
1) Re-weighting on the time steps based on heuristic rules. 
P2~\citep{choi2022perception} and Min-SNR~\citep{hang2023efficient} use monotonous and single-peak weighting strategies according to sign-to-noise ratios (SNR) in different time steps.
2) Re-sampling the time steps. Log-Normal~\citep{karras2022elucidating} assigns high sampling probabilities for the middle time steps of the diffusion process.
CLTS~\citep{xu2024towards} proposes a curriculum learning based time step schedule, gradually tuning the sampling probability from uniform to Gaussian by interpolation for acceleration as shown in Fig.~\ref{fig:1b}.

To investigate the essence of the above accelerations, we take a closer look at the time steps.
\begin{figure}
    \centering
    \includegraphics[width=0.5\textwidth]{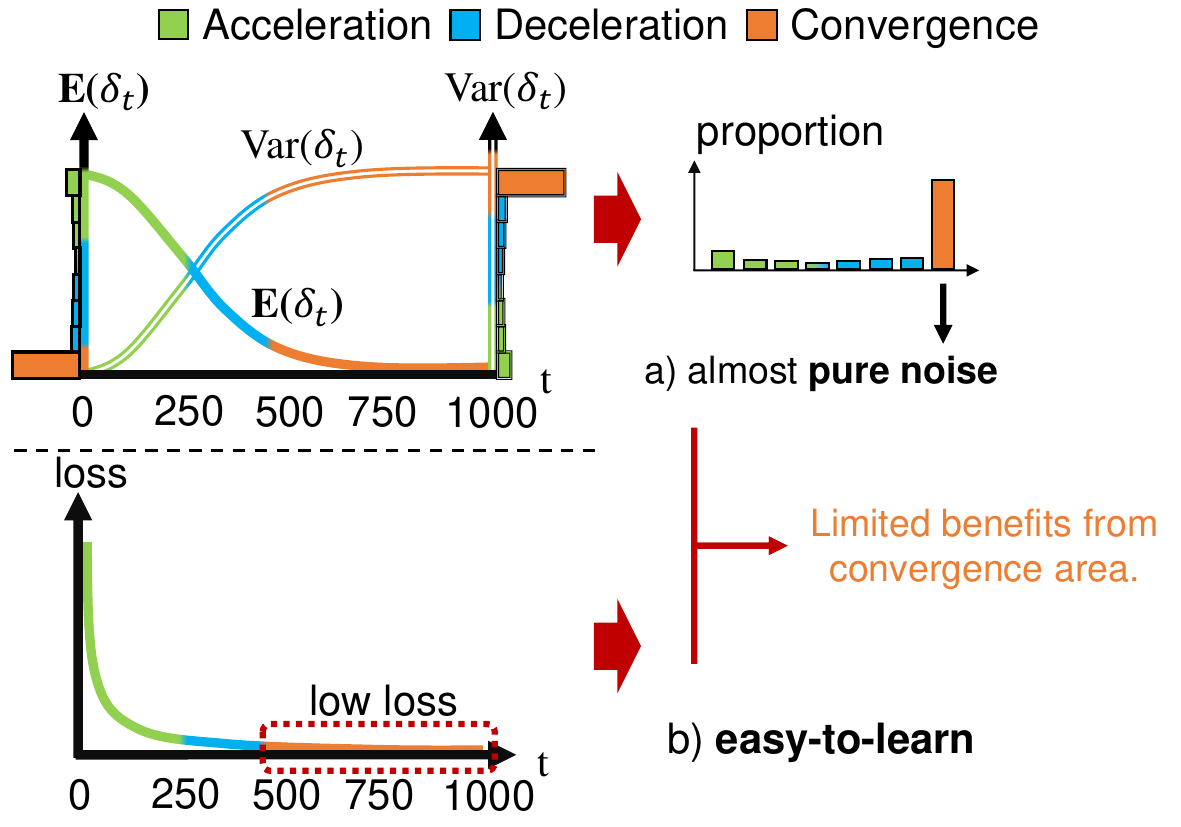}
    \caption{Closer look at time steps: More than half of the time steps are almost pure noise and easy-to-learn. Motivation: designing an efficient training via analyzing process increment $\delta_{t}$ at different time steps. $\mathbf{E}(\delta_{t})$ and $\text{Var}(\delta_{t})$ are the mean and variance of process increments $\delta_{t}$.
Two histograms represent the proportions of the process increments at different noise levels (left) and the proportions of the time steps (right) in the three areas. The loss curve is obtained from DDPM~\citep{ho2020denoising} on CIFAR-10~\citep{krizhevsky2009learning}.
}
\vspace{-20pt}
    \label{fig:1a}
\end{figure}
As shown in the left of Fig.~\ref{fig:1a}, we visualize the changes of mean and variance of the \textit{process increment} $\delta_{t}:=x_{t+1}-x_{t}$ at time step $t$. The time steps are divided into three areas: \textcolor{AccelerationColor}{acceleration}, \textcolor{DecelerationColor}{deceleration}, and \textcolor{ConvergenceColor}{convergence}.
We identify the characteristics of three areas of the time steps.  One can easily find that the proportions of the three areas are \textit{imbalanced: a much larger number of time steps at the narrow convergence area}. Besides, process increments at the convergence area are almost identical noise, \textit{e.g.}, in DDPM, the distribution are
almost pure white noise.
To further explore the characteristics of these three areas, we visualize the training loss curve in the right of Fig.~\ref{fig:1a}.
The loss values from the convergence area are much lower than the others, which indicates estimating the identical noise is easy.

Previous acceleration works have achieved promising results, but the analysis of time steps remains relatively under-explored. P2~\citep{choi2022perception} and Min-SNR~\citep{hang2023efficient} are two re-weighting methods, with their weighting curves across time steps as shown in Fig.~\ref{fig:1b}. They employ uniform sampling of time steps, which include too many easy samples from the convergence area during diffusion model training.
Most re-sampling methods heuristically emphasize sampling the middle-time steps, but they do not dive into the difference between the acceleration and convergence areas.
CLTS~\citep{xu2024towards} gradually changes the sampling distribution from uniform to Gaussian by interpolation as shown in Fig.~\ref{fig:1b}.
The sampling probability of the acceleration area drops faster than the one of the convergence area.
The acceleration area is still under-sampled and therefore not well-learned.

\begin{figure}[t]
\centering
\vspace{-20pt}
\includegraphics[width=\linewidth]{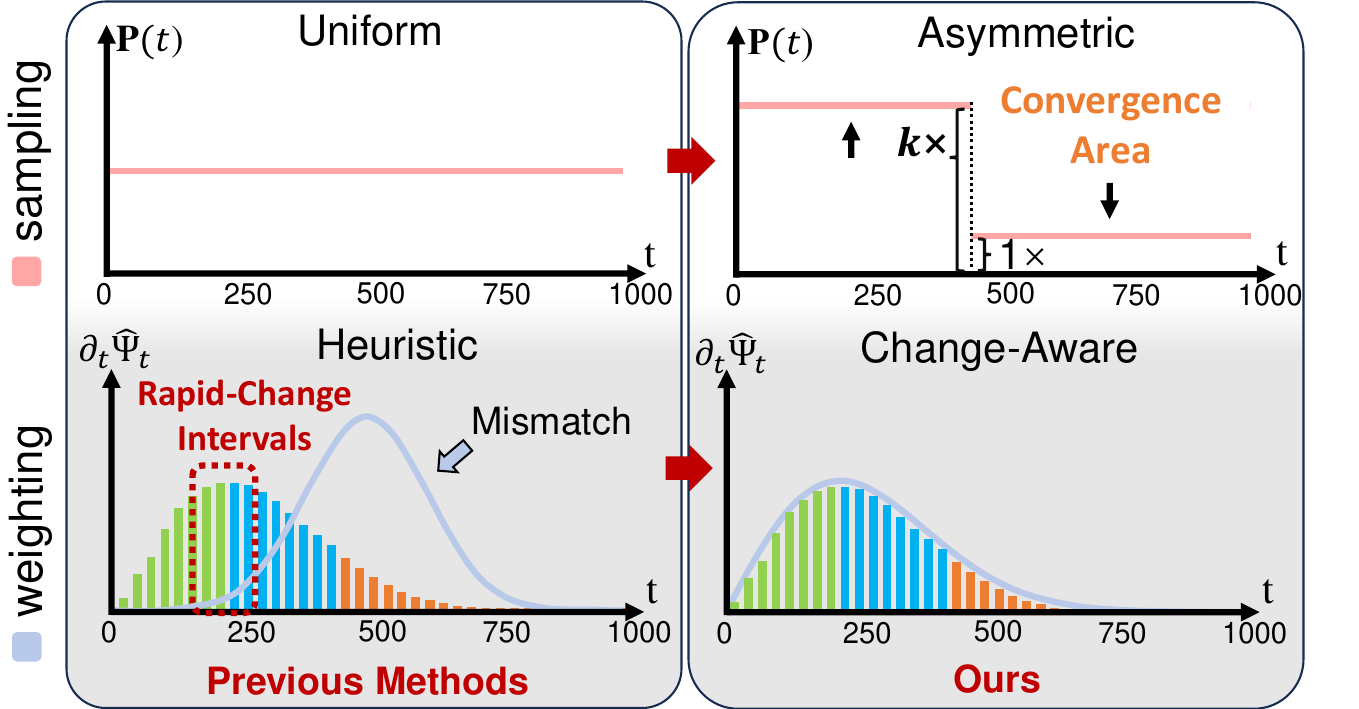}
\caption{Core designs of SpeeD.
Red and blue lines denote sampling and weighting curves.
}
\vspace{-10pt}
\label{fig:method_intro}
\end{figure} %

Motivated by the
analyses from a closer look at time steps, we propose \textit{\methodname}, a novel approach that aims to
improve the training efficiency for diffusion models.
The core ideas are illustrated in Fig.~\ref{fig:method_intro}.
To mitigate the redundant training cost, different from uniform sampling, we design an asymmetric sampling strategy that suppresses the attendance of the time steps from the convergence area in each iteration.
Meanwhile, we weight the time steps by the change rate of the process increment, emphasizing the importance of the rapid-change intervals.

Our approach has the following characteristics:
\methodname is compatible with various diffusion model training methods, \textit{i.e.}, U-Net~\citep{ronneberger2015u} and DiT~\citep{peebles2023scalable} with minimal modifications.
For performance, \methodname achieves non-trivial improvements than baseline and other methods at the same training iterations. 
For efficiency, \methodname consistently accelerates the diffusion training by 3$\times$ across various tasks and datasets. It helps mitigate the heavy computational cost for diffusion model training,
enabling more researchers to train a model at an acceptable expense.
The extra time complexity of \methodname is $\mathcal{O}(1)$,
costing only seconds to reduce days of diffusion models training on datasets like FFHQ, MetFaces and ImageNet-1K. We hope this work can bring novel insights for efficient diffusion model training.

\section{Speeding Up Training: Time Steps}

In this section, we first introduce the preliminaries of diffusion models and then focus on a closer look at time steps and key designs of our proposed \methodname.

\subsection{Preliminaries of Diffusion Models}

In conventional DDPM~\citep{ho2020denoising, sohl2015deep}, given data $x_{0} \sim p(x_0)$, the forward process is a Markov-Gaussian process that gradually adds noise to obtain a sequence \{$x_1, x_2, . . . , x_T$\},
\begin{equation*}
\begin{aligned}
    q(x_{t}|x_{t-1}) &= \mathcal{N}(x_{t}; \sqrt{1-\beta_{t}}x_{t-1}, \beta_{t}\mathrm{\textbf{I}}), \\
    q(x_{1:T}|x_{0}) &= \prod_{t=1}^{T}q(x_{t}|x_{t-1}),
\end{aligned}
\end{equation*}
where $\mathbf{I}$ is the unit matrix, $T$ is the total number of time steps, $q$ and $\mathcal{N}$ represent forward process and Gaussian dist. parameterized by scheduled hyper-parameters $\{\beta_{t}\}_{t\in[T]}$. Perturbed samples are sampled by $x_t=\sqrt{\bar\alpha_t}\cdot x_0+\sqrt{1-\bar\alpha_t}\cdot\epsilon, \epsilon\sim\mathcal{N}(0,\mathbf{I})$,
where $\alpha_t = 1 - \beta_t $ and $\bar{\alpha}_{t}=\prod_{s=1}^{t}\alpha_{s}$.

For diffusion model training, the forward process is divided into  pairs of samples and targeted process increments by time steps $t$, defined as $\delta_{t}:=x_{t+1}-x_{t}$. The diffusion model is expected to predict the next step from the given time step.
The training loss~\citep{ho2020denoising} for diffusion models is to predict the normalized noise. The loss highlighted with weighting and sampling modules, where $\epsilon \sim \mathcal{N}(0,\mathbf{I})$:
\begin{equation}
\begin{aligned}   
L
=\mathbb{E}_{\mu_{t}}[w_{t}||\epsilon-\epsilon_\theta(x_t,t)||^2]
:=\int_{t}w_{t}||\epsilon-\epsilon_{\theta}(x_{t},t)||^{2}\mathbf{d}\mu_{t},
\end{aligned}
\label{eqn:DDPM_Obj}
\end{equation}
Intuitively, a neural network $\epsilon_{\theta}$  is trained to predict the normalized noise $\epsilon$ added at given time-step $t$. The probability of a sample being sampled in the forward process is determined by the probability measure $\mu_{t}$, while the weight of the loss function is determined by $w_{t}$ at $t^{\text{th}}$ time-step.

\subsection{Overview of \methodname} 

Based on the above observations and analyses, we propose \methodname, a novel approach for achieving lossless training {acceleration} tailored for diffusion models.
As illustrated in Fig. 
\ref{fig:method_intro}, 
\methodname suppresses the trivial time steps from {convergence} area, and weight the rapid-change intervals between {acceleration} and {deceleration} areas. 
Correspondingly, two main modules, asymmetric sampling and change-aware weighting, are proposed. Asymmetric sampling uses a two-step step function to respectively suppress and increase the sampling probability corresponding to trivial and beneficial time steps. Change-aware weighting is based on the change rate of process increment $\partial_{t}\Psi(t)$ in Theorem~\ref{theo:bound}.

\subsection{Asymmetric Sampling}
\label{ssec:asymmetric_sampling}
\methodname adopts the time steps sampling probability $\mathbf{P}(t)$ as the step function in Eqn.~\ref{eqn:sampling} to construct the loss in Eqn.~\ref{eqn:DDPM_Obj}. We first define $\tau$ as the step threshold in $\mathbf{P}(t)$.
The pre-defined boundary $\tau$ means the area where the time step are suppressed. The sampling probability is $k$ times from time-steps where $t < \tau$ than that from $t > \tau$ instead of the uniform sampling $\mathbf{U}(t) = 1 / T$.

\vspace{-5pt}
\begin{equation}
\label{eqn:sampling}
    \mathbf{P}(t)= \begin{cases}
        \frac{k}{T+\tau(k-1)}
       ,\quad & 0 < t \leq \tau; \\
       \frac{1}{T+\tau(k-1)},\quad &  \tau < t \leq T,
        \end{cases}       
\end{equation}
where suppression intensity $k \geq 1 \text{ and } \tau \in (0,T]$.

\paragraph{Threshold Selection $\tau$.} According to Theorem~\ref{theo:bound}, given a magnitude $r$, $\tau$ should satisfy $\hat{r}(\tau) > r$ to make sure that ${\tau} > t_{\text{d-c}}$, where the time steps suppressed are all time steps in the {convergence} area. To maximum the number of suppressed time steps, we set $\tau\leftarrow \sqrt{2T\log r/\Delta_{\beta}+T^{2}\beta_{0}^{2}/\Delta_{\beta}^{2}}-T\beta_{0}/\Delta_{\beta}$.

\subsection{Change-Aware Weighting}
\label{ssec:rapid-aware_weighting}
According to Theorem~\ref{theo:bound}, a faster change of process increment means fewer samples at the corresponding noise level.
This leads to under-sampling in {acceleration} and deceleration areas.
Change-aware weighting is adopted to mitigate the under-sampling issue.
The weights $\{w_{t}\}_{t\in[T]}$ are assigned based on the gradient of the variance over time, where we use the approximation $\partial_{t}\hat{\Psi}_{t}$ in Theorem~\ref{theo:bound}.

The original gradient $\partial_{t}\hat{\Psi}_{t}$ is practically not suitable for weighting due to its small scale. Therefore, $\partial_{t}\hat{\Psi}_{t}$ is re-scaled into range $[1\text{-}\lambda,\lambda]$ that $\min\{1,\max_{t}{\partial_{t}\hat{\Psi}_{t}}\} \rightarrow \lambda$ and $\max\{0,\min_{t}{\partial_{t}\hat{\Psi}_{t}}\} \rightarrow 1\text{-}\lambda$, where symmetry ceiling $\lambda  \in [0.5, 1]$. $\lambda$ regulates the curvature of the weighting function. A higher $\lambda$ results in a more obvious distinction in weights between different time-steps.

\subsection{Case Study: DDPM} 
\label{ssec:closer_look}
In DDPM, %
the diffusion model learns the noise added in the forward process at given $t^{\text{th}}$ time step. The noise is presented as $\epsilon$, the label in Eqn.~\ref{eqn:DDPM_Obj}, which is the normalized process increment at given time step. This label tells what the output of the diffusion model is aligning to. To take a closer look, we focus on the nature of the  \deltaname $\delta_{t}$ itself to study the diffusion process $x_{t}\rightarrow x_{t+1}$, instead of $\epsilon$ the normalized one. According to Theorem~\ref{theo:bound} and Remark~\ref{rmk:bounds}, based on the variation trends of process increments $\delta_{t}$, we can distinguish three distinct areas: \textcolor{AccelerationColor}{acceleration}, \textcolor{DecelerationColor}{deceleration}, and \textcolor{ConvergenceColor}{convergence}.

\begin{theorem}[\Deltaname in DDPM]
\label{theo:bound}
    In DDPM's setting~\citep{ho2020denoising}, the linear schedule hyper-parameters $\{\beta_{t}\}_{t\in[T]}$ is an equivariant series, the extreme deviation $\Delta_{\beta} := \max_{t} \beta_{t}-\min_{t}\beta_{t}$, $T$ is the total number of time steps, and we have the bounds about the \deltaname $\delta_{t}\sim\mathcal{N}(\phi_{t},\Psi_{t})$, where $
    \phi_{t}:=(\sqrt{\alpha_{t+1}}-1)\sqrt{\bar{\alpha}_{t}}x_{0},
    \Psi_{t}:=
    [2-\bar{\alpha}_{t}(1+\alpha_{t+1})]\mathbf{I}
    $, $\mathbf{I}$ is the unit matrix:    
    \vspace{-5pt}
    \begin{equation}
    \begin{aligned}
    & \colorbox{red!30}{\text{Upper-bound: }} ||\phi_{t}||^{2} \leq \hat{\phi}_{t}||\mathbb{E}x_{0}||^{2},
    \\
    & \colorbox{blue!10}{\text{Lower-bound: }}
     \Psi_{t}
    \succeq\hat{\Psi}_{t} \mathbf{I}, 
    \end{aligned}
    \vspace{-10pt}
    \end{equation}
    where $~\hat{\phi}_{t}:=\beta_{\max}\exp\{-(\beta_{0}+{\Delta_{\beta}t}/{2T})t\}$ and $~\hat{\Psi}_{t}:= 2-2\exp\{-(\beta_{0}+{\Delta_{\beta}t}/{2T})t\}$.
\end{theorem}

\begin{figure*}[h]
\centering
\subfloat[\colorbox{red!30}{Upper-bound's factor}: $\hat{\phi}_{t}$.]{
\includegraphics[width=0.32\linewidth]{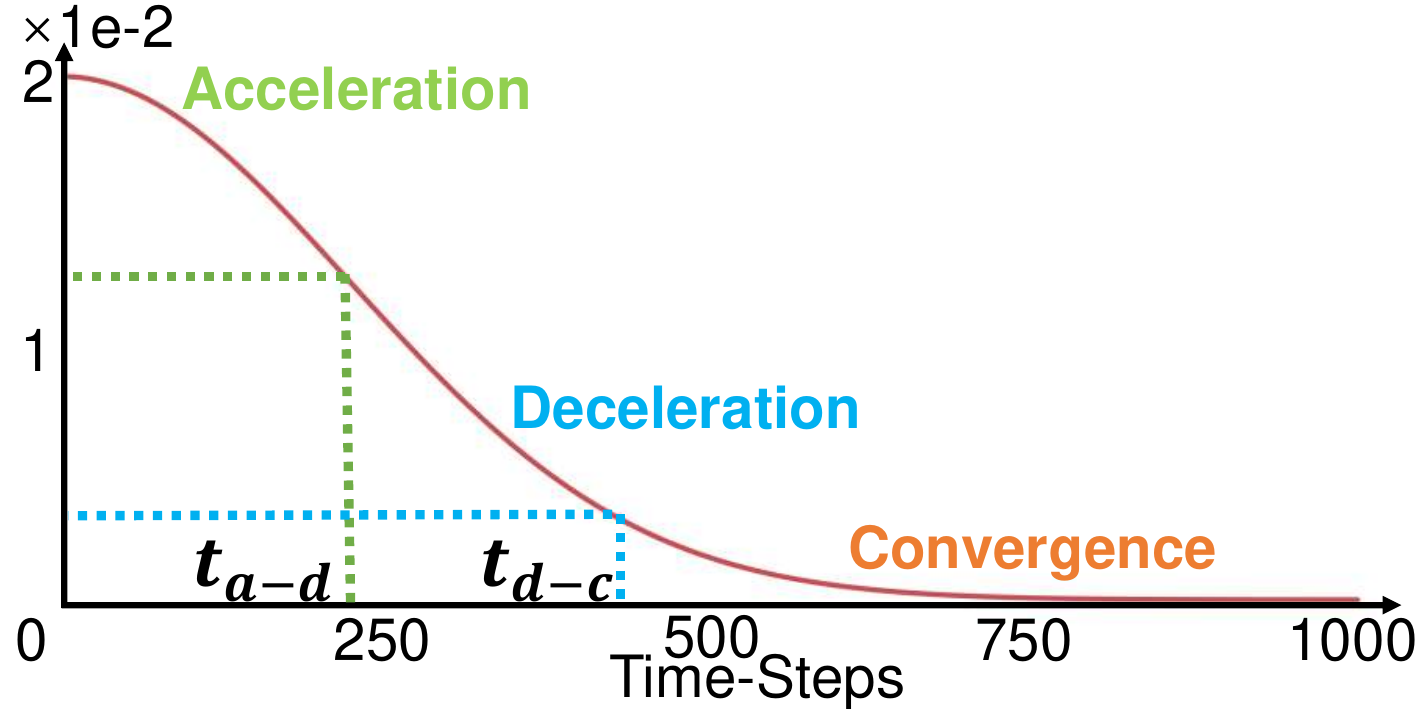}
}
\subfloat[\colorbox{blue!10}{Lower-bound's factor}: $\hat{\Psi}_{t}$.]{
\includegraphics[width=0.32\linewidth]{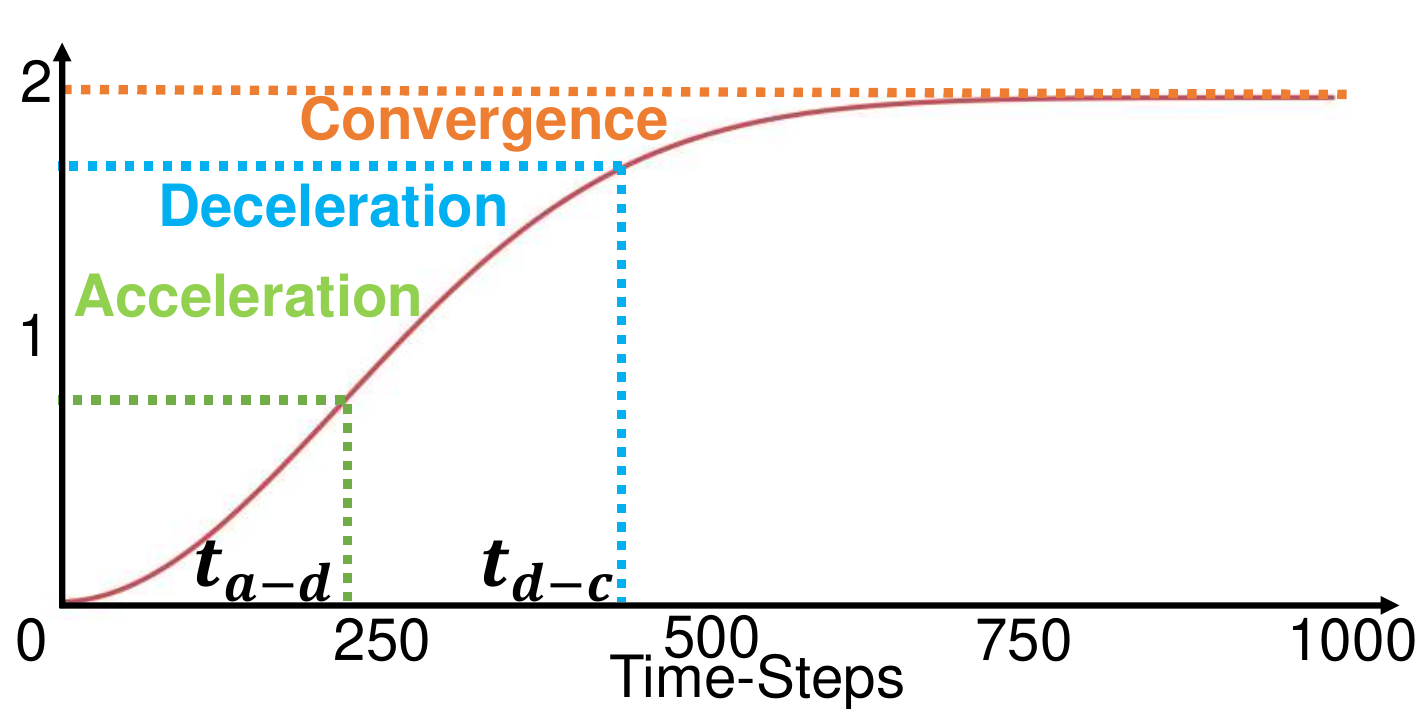}
}
\subfloat[Partial derivative: $\partial_{t}\hat{\Psi}_{t}$.]{
\includegraphics[width=0.32\linewidth]{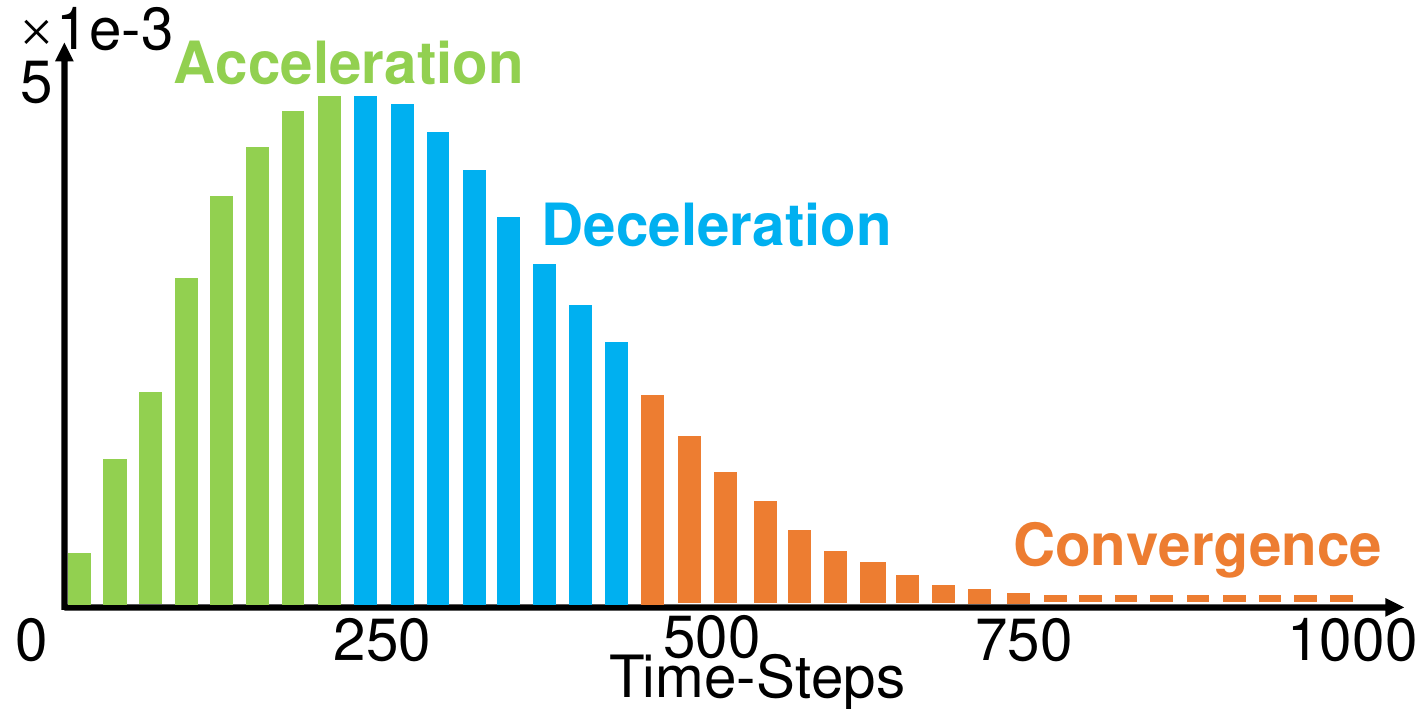}
}
\caption{Visualization of Theorem~\ref{theo:bound}: three areas of acceleration, deceleration and convergence.} 
\label{fig:bounds}
\vspace{-10pt}
\end{figure*}

\begin{remark}
\label{rmk:bounds}
    The entire diffusion process can be approximated using the upper and lower bounds from Theorem~\ref{theo:bound}, which we visualize as shown in Figure~\ref{fig:bounds}. We can observe that the diffusion process can be divided into three areas: acceleration, deceleration, and convergence. The two boundary points of these areas are denoted as \(t_{\text{a-d}}\) and \(t_{\text{d-c}}\) with their specific definitions and properties outlined below.
\end{remark}

\textbf{Definition of $t_{\text{a-d}}$.} The boundary between the \textcolor{AccelerationColor}{acceleration} and \textcolor{DecelerationColor}{deceleration} areas is determined by the inflection point in the parameter variation curves, as illustrated in Figure~\ref{fig:bounds}. This inflection point represents the peak where the process increment changes most rapidly. The key time-step $t_{\text{a-d}}$ between {acceleration} and {deceleration} areas satisfies $t_{\text{a-d}} = \arg\max_{t}\partial_{t}\hat{\Psi}_{t}$ and $\beta_{t_{a-d}} = \sqrt{{\Delta_{\beta}} /{T}}$ in our setting, where $\partial_{t}\hat{\Psi}_{t} = 2(\beta_{0}+{\Delta_{\beta}t}/{T})\exp\{-(\beta_{0}+{\Delta_{\beta}t}/{2T})t\}$.

\textbf{Definition of $t_{\text{d-c}}$.} The process is considered to be in the \textcolor{ConvergenceColor}{convergence} area where the increments' variance is within a range. The convergence area is identified by a magnitude $r$, where $1-1/r$ is the ratio to the maximum variance.

According to Theorem~\ref{theo:bound}, the {convergence} area is defined as \textit{one magnitude reduction} of the scale factor (\textit{i.e.}, $1\times r$), and we have the lower-bound of the magnitude $\hat{r}_{t}:=\exp\{(\beta_{0}+\Delta_{\beta}t/2T)t\}$ employed as the threshold selection function in Section~\ref{ssec:asymmetric_sampling}. The time step $t$ is guaranteed to be in {convergence} area when $\hat{r}_{t} > r$, .

\paragraph{Analyses.}
In the {convergence} area, the variations of $\delta_{t}$ stabilize, signifying that the process is getting steady. 

This area corresponds to a very large proportion of the overall time steps. On top of that, the training loss in this area is empirically low, which leads to the redundant training cost on time steps with limited benefits for training.

In the {acceleration} area, the variations of $\delta_{t}$ increase, indicating a rapid change. Conversely, in the {deceleration} area, the variations of $\delta_{t}$ decrease, reflecting a slowing down of the process. Notably, near the peak between the {acceleration} and {deceleration} areas, the process exhibits the fastest changes.
These time steps only occupy a small proportion.
Beyond that, the training losses in this area are empirically high.
The issue is that a hard-to-learn area is even under-sampled, necessitating more sampling and training efforts.
\paragraph{Takeaways.}
Based on the analyses and observations, we provide takeaways as follows:
    \begin{itemize}
        \item The samples from {convergence} area provide limited benefits for training. The sampling of the time step from this area should be suppressed.
        
        \item Pay more attention to the process increment’s rapid-change area which is hard to learn and the corresponding time steps are fewer than the other areas.
    \end{itemize}

\subsection{General Cases: Beyond DDPM}
\label{ssec:general}

This section generalize the above findings on DDPM to broader settings. The findings are about the process increments $\delta_{t}:=x_{t+1}-x_{t}$, and the related differentiation of right limit $\mathbf{d} x$ in forward process.
The corresponding SDE~\citep{karras2022elucidating} and the discretization are:
$$\mathbf{d} x = x\dot{s} /s \mathbf{d} t + s \sqrt{\dot{\sigma}\sigma}\mathbf{d} w,~ x_{t}=s_{t}x_{0}+s_{t}\sigma_{t}\epsilon, \epsilon \sim \mathcal{N}(0,\mathbf{I}),$$
where scale factor $s=s_{t}$ and noise standard deviation (std.) $\sigma=\sigma_{t}$ are the main designs related to the factors $\hat{\phi}_{t}$ and $\hat{\Psi}_{t}$ about process increment $\delta_{t}$ in Theorem~\ref{theo:bound} at time steps $t$.

\paragraph{Generalize Theorem~\ref{theo:bound}}: $s$-$\sigma$ Scheduled Process Increments. The generalized process increment is $\delta\sim\mathcal{N}(\Delta x_{0}, \Sigma\mathbf{I})$, where $\Delta:=s_{+}-s$ and $\Sigma:=s_{+}^{2}\sigma_{+}^{2}+s^{2}\sigma^{2}$ across $t$. $\Delta$, $\Sigma$ are continuous on $t$ without discretization, where $s_{+}$ and $\sigma_{+}$ are the right outer limits, \textit{i.e.}, $s(t+\mathbf{d} t)$. In discretization, $\Delta_{t}$ and $\Sigma_{t}$, marked by $t$, are related to sample granularity of time step $t$. Like Theorem~\ref{theo:bound}, we study the variation of process increments by $\dot{\Delta}=\dot{s}_{+}-\dot{s}$ and 
$\dot{\Sigma}=\mathbf{m}^{\top}\dot{\mathbf{n}}$
where $\mathbf{m}=[s_{+}^{2},\sigma_{+}^{2},s^{2},\sigma^{2}]^{\top}$, $\mathbf{n}=[\sigma_{+}^{2},s_{+}^{2},\sigma^{2},s^{2}]^{\top}$. This formulation involves only terms about derivatives of given schedule functions, which brings computational convenience. Tab.~\ref{tab:ingredients} (in Appendix) provides all ingredients needed to calculate curves of $\dot{\Delta}$ and $\dot{\Sigma}$ in schedules of VP, VE and EDM.
We also generalize the previous \textit{takeaways} from the DDPM case study to $s$-$\sigma$ scheduled setting with the following analyses.

\textit{$\sigma$ is better for the design of sampling and weighting than $s$}. It stands due to its direct reflection about SNR (signal-to-noise ratio), and additionally, because $s$ is usually adapted to heuristic motivations. In DDPM, corresponding to VP in Tab.~\ref{tab:ingredients}, the SDE design is simply from data to normal noise. In VE, realistic diffusion processes inspire that the diffusion rate is limited to $\sigma=\sqrt{t}$. Further in EDM, motivation become more complex of training objective and concise of schedule and motivation at the same time, bringing benefits to training. Its key ideas and designs are 1) the std. of inputs and targeted outputs of a neural network $F_{\theta}$ in EDM is constrained to 1 with preconditioning; 2) the weights $w_{t}$ in Eqn.~\ref{eqn:DDPM_Obj} are allocated according to $c_{\text{out}}w_{t}=1$, where $c_{\text{out}}$ is the scale factor of F-prediction neural network $F_{\theta}$'s outputs, and is related to $\sigma$ and the std. of data (sometimes normalized as 1).

\textit{Sampling deserves more attention.} The SDE design goal of most diffusion models nowadays is to add much larger noise to the data so that the samples can cover larger space. However, in terms of the process increment, it always results in a low signal-to-noise ratio of the late data when $t$ is large. Either the standard deviation is too large or the $s$ used to suppress it is too small. For instance, EDM does not bias the data distribution from expectations due to the scale $\dot{\Delta}=0$, but $\dot{\Sigma}=2[(t+\mathbf{d}t)^2+t^2]$ is a quadratic increase as $t$ grows. In VP, $s\rightarrow 0$, as $t$ grows, leads that the model needs to recover expectations in approximation. For these samples, which are not very informative, a single weight adjustment is not as efficient as reducing the sampling rate.

\section{Experiments}
In this section, the visualization is provided in Section~\ref{vis_main_paper}. Comparison with SOTA methods in is provided mainly on ImageNet-1K dataset. The compared baselines and scheduler settings including SDE-VP/VE and EDM. The comparison and ablations on most related hyperparameters and tasks are empirically verified.

\subsection{Visualization}
\label{vis_main_paper}

\begin{figure*}[t]
    \centering
    
    \begin{subfigure}{0.48\linewidth}
        \includegraphics[width=1\linewidth]{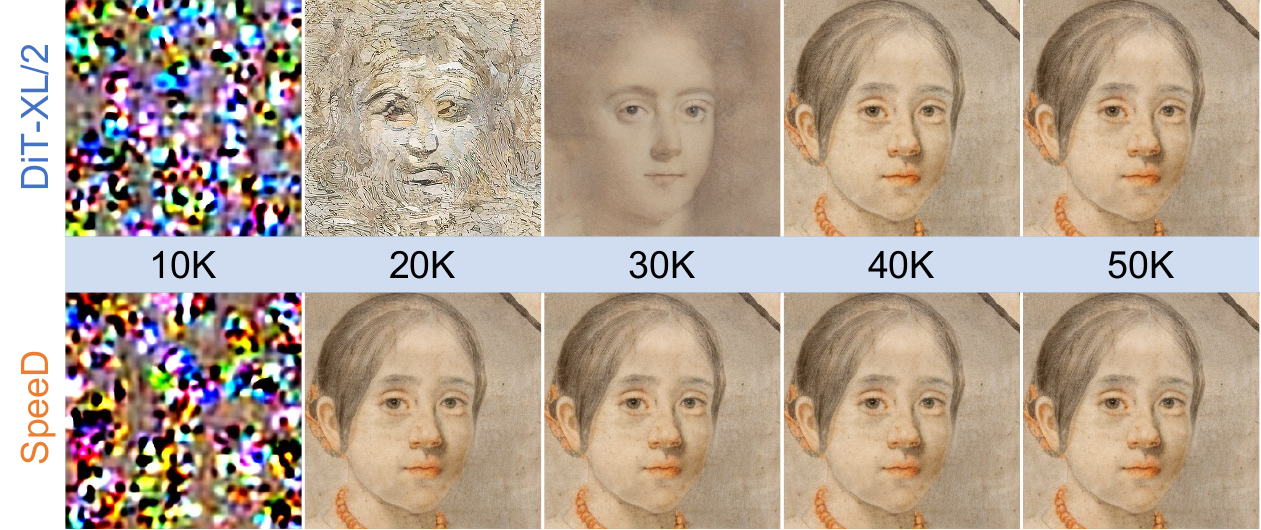}
        \caption{Visualization comparison on MetFaces dataset.}
        \label{fig:vis_metface}
    \end{subfigure}
    \begin{subfigure}{0.48\linewidth}
        \includegraphics[width=1\linewidth]{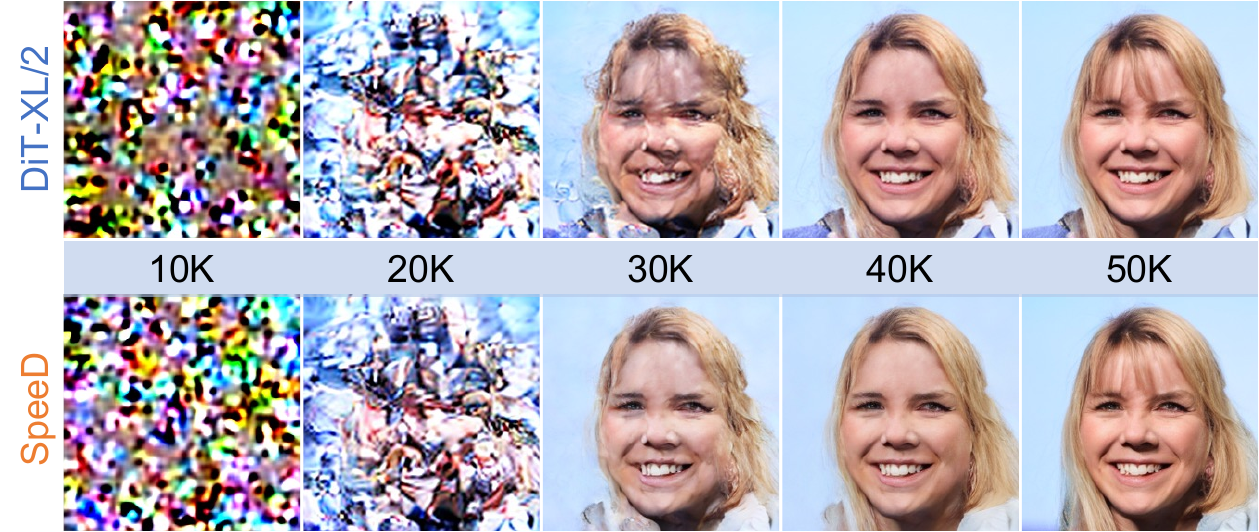}
        \caption{Visualization comparison on FFHQ dataset.}
        \label{fig:vis_ffhq}
    \end{subfigure}
    \caption{Our SpeeD obtains significant improvements than baseline in visualizations. More visualizations on other datasets and tasks can be found in the Appendix~\ref{vis_appendix}.}
    \label{fig:vis_vs_baseline}
    \vspace{-10pt}
\end{figure*}

The comparison of visualizations between SpeeD and DiT-XL/2 models on the MetFaces and FFHQ datasets clearly demonstrates the superiority of SpeeD. As shown in Fig.~\ref{fig:vis_vs_baseline}, SpeeD achieves significantly better visual quality at just 20K or 30K training iterations, compared to DiT-XL/2. This highlights that SpeeD reaches high-quality results much faster than the baseline method, making it a more efficient and effective approach for training diffusion models.

\subsection{Implementation Details}
\label{sec:implementation}

\textbf{Datasets.} We mainly investigate the effectiveness of our approach on the following datasets: MetFaces~\citep{karras2020training} and FFHQ~\citep{karras2019style} are used for unconditional tasks, Celeb-A~\citep{liu2015deep}, CIFAR-10~\citep{krizhevsky2009learning} and ImageNet-1K~\citep{deng2009imagenet} are used to train conditional image generation, and MS-COCO~\citep{lin2014microsoft} is used to evaluate the generalization of our method in the text to image task. More details can be found in the Appendix~\ref{details_of_training_appendix}.

\textbf{Network architectures.} U-Net~\citep{ronneberger2015u} and DiT~\citep{peebles2023scalable} are two famous architectures in the diffusion model area. We implement our approach on these two architectures and their variants. We follow the same hyper parameters as the baseline by default.
More information about the details of the architectures can be found in Appendix \ref{arch_details}.

\textbf{Training details.}
We train all models using AdamW ~\citep{kingma2014adam, loshchilov2017decoupled} with a constant learning rate 1e-4. We set the maximum step in training to 1000 and use the linear variance. All images are augmented with horizontal flip transformations if not stated otherwise. Following common practice in the generative modeling literature, the exponential moving average (EMA)~\citep{gardner1985exponential} of network weights is used with a decay of 0.9999. 
The results are reported using the EMA model. 
Details can be found in Tab.~\ref{tab:arch_conf}.

\textbf{Evaluation protocols.} 
In inference, we default to generating 10K images.
Fréchet Inception Distance (FID) is to evaluate both the fidelity and coverage of generated images.

\subsection{Comparisons with other strategies.}
\label{sec:comparison}

\newcommand{\blue}[1]{$_{\color{RoyalBlue}\downarrow #1}$}
\newcommand{\upred}[1]{$_{\color{RedOrange }}\downarrow$}
\begin{table*}[tp]
 
    \centering
    \footnotesize
    \tablestyle{12.2pt}{1.3}
    \begin{tabular}{cccccccccccc}
    \multicolumn{2}{c}{dataset}  & \multicolumn{5}{c}{MetFaces} & \multicolumn{5}{c}{FFHQ}  \\
    \multicolumn{2}{c}{iterations}  & 10K & 20K & 30K & 40K & 50K & 10K & 20K & 30K & 40K & 50K \\ 
    \shline

     \multicolumn{2}{c}{\baseline{DiT-XL/2~\citep{peebles2023scalable}}}
     & 
     \baseline{398.7}
     & \baseline{132.7}
     & \baseline{74.7}
     & \baseline{36.7}
     & \baseline{29.3}
     & \baseline{356.1}
     & \baseline{335.3}
     & \baseline{165.2}
     & \baseline{35.8}
     & \baseline{12.9}
     
    \\
    \multicolumn{2}{c}{P2~\citep{choi2022perception}} &
     377.1 & 328.2 & 111.0 & 27.3 & 23.3 & 368.7 & 357.9 & 158.6 & 35.5 & 15.0 
     \\
     \multicolumn{2}{c}{Min-SNR~\citep{hang2023efficient}} & 
     389.4 & 313.9 & 52.1 & 31.3 & 28.6 & 376.3 & 334.1 & 151.2 & 34.0 & 12.6 
     \\
     \multicolumn{2}{c}{Log-Normal~\citep{esser2024scaling}} &
     \textbf{311.8} & 165.1 & 63.9 & 51.1 & 47.3 & \textbf{307.6} &  165.1 & \textbf{63.9} & 51.1 & 47.3 
     \\
    \multicolumn{2}{c}{CLTS~\citep{xu2024towards}} &    
    375.0 & 57.2 & 28.6 & 24.6 & 23.5 & 336.1 & 329.1 & 173.4 & 33.7 & 12.7
     \\
     \multicolumn{2}{c}{\default{SpeeD (ours)}}
     
     & \default{367.3}
     & 
     \default{\textbf{23.4}}
     & 
     \default{\textbf{22.6}}
     & 
     \default{\textbf{22.1}}
     & 
     \default{\textbf{21.1}}
     & 
     \default{322.1}
     &
     \default{\textbf{320.0}}
     &
     \default{91.8}
     &
     \default{\textbf{19.8}}
     & 
     \default{\textbf{9.9}}
     
     \\

    \end{tabular}
    \caption{The FID$\downarrow$ comparison to the baseline: DiT-XL/2, re-weighting methods: P2 and Min-SNR, and re-sampling methods: Log-Normal and CLTS. All methods are trained with DiT-XL/2 for 50K iterations. We report the FID per 10K iterations. Our approach achieves the best results on MetFaces and FFHQ datasets.\textbf{ Bold entries} are best results. Following previous work~\citep{go2023addressing}, more results of 100K iterations and \textit{longer training phase} with different schedules are in Appendix~\ref{appdx_add_exp}.
    }
\label{tab:compare_sota}
\vspace{-10pt}
\end{table*}
\paragraph{Performance comparisons.} 
Before our comparison, we first introduce our baseline, \textit{i.e.}, DiT-XL/2, a strong image generation backbone as introduced in DiT~\citep{peebles2023scalable}. We follow the hyperparameter settings from DiT and train DiT-XL/2 on MetFaces~\citep{karras2020training} and FFHQ~\citep{karras2019style}, respectively. 
We compare our approach with two re-weighting methods: P2~\citep{choi2022perception} and Min-SNR~\citep{hang2023efficient}, and two re-sampling methods: Log-Normal~\citep{karras2022elucidating} and CLTS~\citep{xu2024towards}. 
In the evaluation, we use 10K generated images to calculate FID~\citep{heusel2017gans} for comparison. 
To make a detailed comparison, all approaches are trained for 50K iterations and we report the FID per 10K iterations.

As shown in Tab~\ref{tab:compare_sota}, compared to DiT-XL/2, re-weighting, and re-sampling methods, our approach obtains the best FID results. Specifically, at the 50K iteration, compared to other methods, we reduce 2.3 and 2.6 FID scores on MetFaces and FFHQ at least. Another interesting finding is that the re-weighting methods reduce the FID slowly at the beginning of the training, \textit{i.e.}, from the 10K to 20K iterations. That aligns with our analysis well: re-weighting methods involve a lot from the convergence area. Based on the experimental results, the time steps from the convergence area indeed contribute limited to training.

\begin{figure*}[t]
    
    \centering
    \begin{subfigure}{0.32\linewidth}
        \includegraphics[width=1\linewidth]{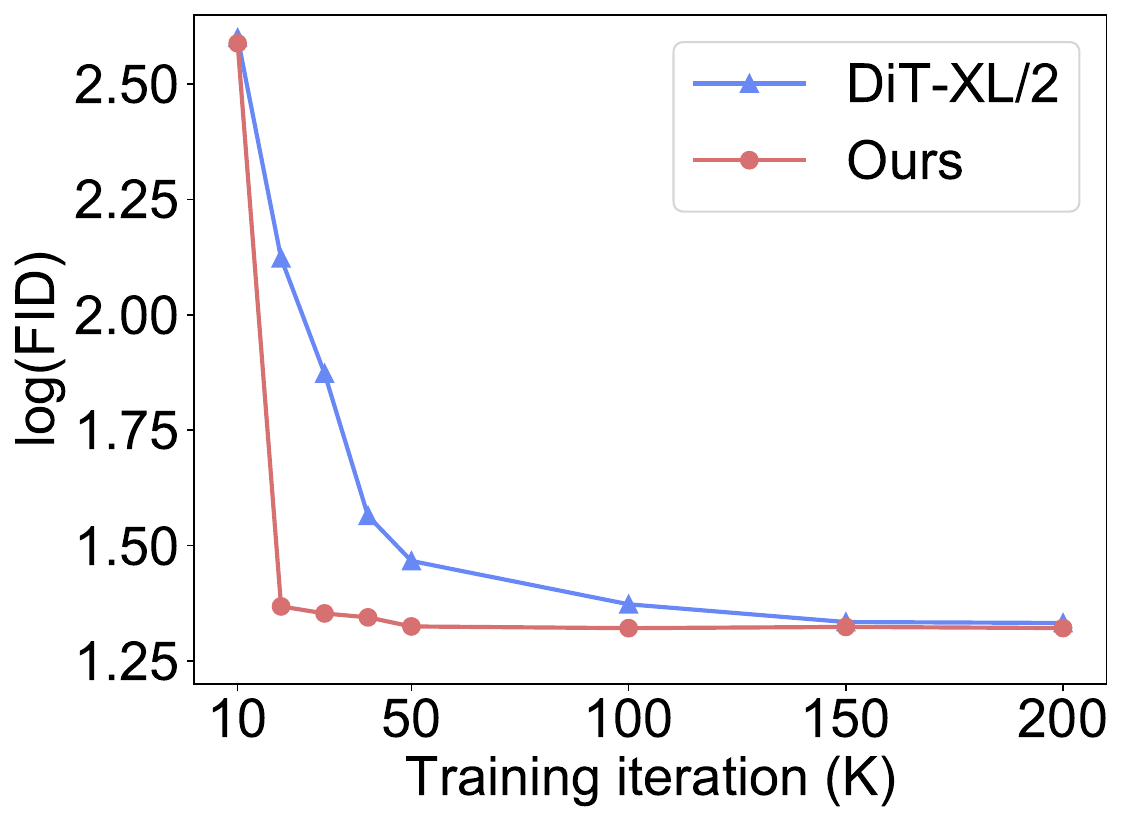}
    \end{subfigure}
     \begin{subfigure}{0.32\linewidth}
        \includegraphics[width=1\linewidth]{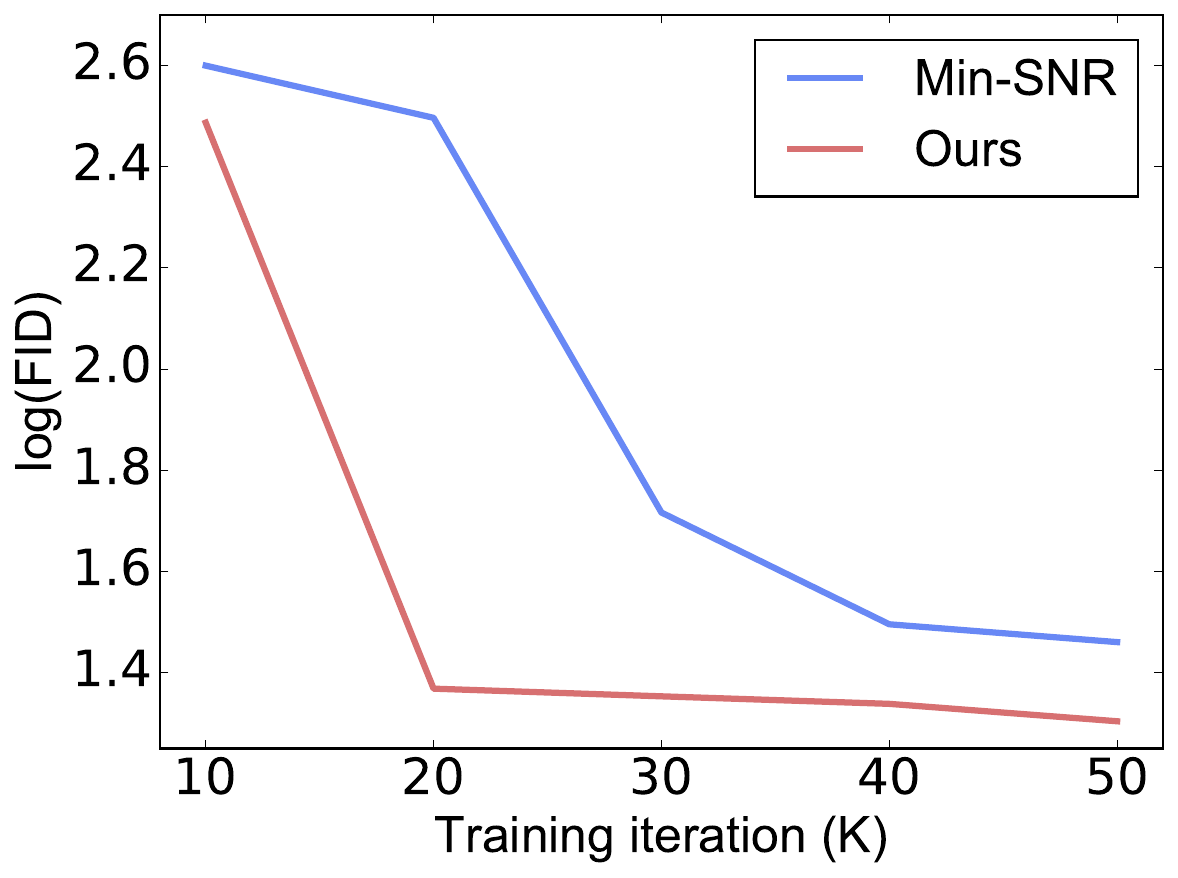}
    \end{subfigure}
     \begin{subfigure}{0.32\linewidth}
        \includegraphics[width=1\linewidth]{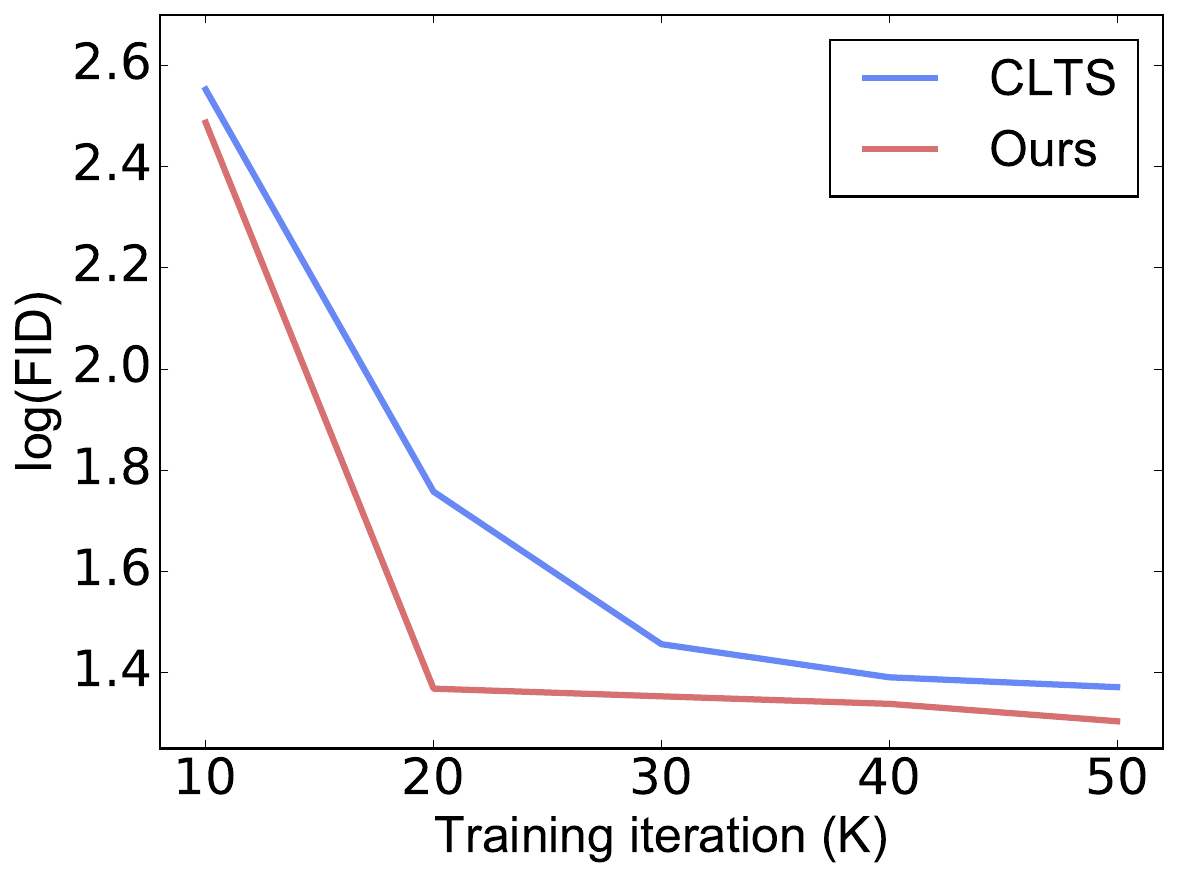}
    \end{subfigure}
    \vspace{-10pt}
    \caption{We plot the FID-Iteration curves of our approach and other methods on the MetFaces dataset. \textit{SpeeD} accelerates other methods obviously. The horizontal and vertical axes represent the training iterations and log(FID), respectively. Detailed ones are in appendix.}

    \label{fig:efficiency}
\end{figure*}

\paragraph{Efficiency comparisons.} 
In addition to the performance comparison, we also present the acceleration results of our \methodname.
This naturally raises a question: \textit{how to calculate the acceleration? }Here, we follow the previous diffusion acceleration methods~\citep{gao2023masked} and other efficient training papers~\citep{qin2023infobatch, xu2024towards}:
visualizing the FID-Iteration curve and reporting the estimated highest acceleration ratio. We mainly compare with DiT-XL/2, one re-weighting method Min-SNR~\citep{choi2022perception}, and one re-sampling method CLTS~\citep{xu2024towards} in Fig~\ref{fig:efficiency}. 
At the same training iterations, our approach achieves significantly better FID scores than other methods. Notably, SpeeD accelerates the Min-SNR, and CLTS by 2.7 and 2.6 times, respectively (More comparisons in Appendix~\ref{more_results_appendix}).

For the comparison with the baseline, \textit{i.e.}, DiT-XL/2, considering the 50K iterations might be too short for converge, we extend the training iterations from 50K to 200K. In the long-term training, we speed up the DiT-XL/2 by 4 times without performance drops. That shows the strong efficiency of our proposed method. Most importantly, we can save \textbf{3$\sim$5} times the overall training cost with very minimal overhead. For instance, we save 48 hours (result obtained by training on 8 A6000 Nvidia GPUs) of training time for DiT-XL/2 with negligible seconds overhead.

\subsection{Generalization Evaluation} 
\label{sec:generalization}

\begin{table}[t]
\centering
\tablestyle{6.7pt}{1.3}
\begin{tabular}{ccccccc}
\multirow{2}{*}{}           & \multicolumn{2}{c}{Metfaces} 
&\multicolumn{2}{c}{FFHQ} & \multicolumn{2}{c}{ImageNet}   \\ 
& DiT & U-Net  &  DiT & U-Net & DiT & U-Net     \\ \shline
\baseline{baseline}          &  \baseline{36.61}  &  \baseline{46.77} & \baseline{12.86} & \baseline{17.37} & \baseline{26.74} & \baseline{45.71} \\ %
\default{\methodname}    &  \default{\textbf{21.13}}  &  \default{\textbf{22.88}}  & \default{\textbf{9.95}} & \default{\textbf{16.52}}  & \default{\textbf{20.63}} & \default{\textbf{37.33}}  \\ %
\shline
improve     &  15.48 &  23.89 & 2.91  &  0.85  & 6.11 & 7.38  \\
                              
\end{tabular}
\vspace{-10pt}
\caption{Cross-architecture robustness evaluation. `Baseline' denotes training diffusion models without acceleration strategy. 'DiT' refers to the DiT-XL/2 network.
All FID scores are obtained by testing 10K generated images.}
\label{tab:cross_arch}
\vspace{-10pt}
\end{table}
\begin{table}[t]
\tablestyle{7.5pt}{1.3}
\begin{tabular}{ccccccc}
                 & \multicolumn{2}{c}{linear}       &  \multicolumn{2}{c}{quadratic} & \multicolumn{2}{c}{cosine}         \\ 
                 & FID $\downarrow$ & IS $\uparrow$ & FID $\downarrow$ & IS $\uparrow$ & FID $\downarrow$  & IS $\uparrow$ \\ \shline
\baseline{baseline}
& \baseline{12.86}
& \baseline{4.21}
& \baseline{11.12}
&  \baseline{4.21}
& \baseline{18.31}
& \baseline{4.10}
\\
\default{SpeeD}    &  \default{\textbf{9.95}} &  \default{\textbf{4.23}} &  \default{\textbf{9.78}}  &  \default{\textbf{4.29}}  & \default{\textbf{17.79}} & \default{\textbf{4.15}}  \\
\shline
improve   &  2.91 & 0.02  & 1.34 &  0.08  & 0.52 & 0.05 \\ 
\end{tabular}
\vspace{-10pt}
\caption{Comparisons of FID and IS scores on FFHQ with different schedules on time steps. We mainly evaluate the generalization of our approach on linear, quadratic, and cosine schedules. We use the vanilla DiT-XL/2 as the baseline.}
\label{tab:cross_schedule}
\vspace{-20pt}
\end{table}
\textbf{Cross-architecture robustness evaluation.} 
There are mainly two architectures in the diffusion models: U-Net~\citep{ronneberger2015u} and DiT~\citep{peebles2023scalable}.
\methodname is not correlated to specific model architecture, thereby it is a model-agnostic approach. We implement our method with DiT-XL/2 and U-Net on MetFaces, FFHQ, and ImageNet-1K, respectively.
\textbf{We default to training the models for 50K iterations on MetFaces and FFHQ, 400K on ImageNet-1K.}%
To ensure a fair comparison, we keep all hyper-parameters the same and report the FID scores at 50K iterations.
As shown in Tab.~\ref{tab:cross_arch}, SpeeD consistently achieve significantly higher performance under all settings, which indicates the strong generality of SpeeD for different architectures and datasets.

\textbf{Cross-schedule robustness evaluation.} 
In the diffusion process, there are various time step schedules, including linear~\citep{ho2020denoising}, quadratic and cosine~\citep{nichol2021improved} schedules.
We verify SpeeD's robustness across these schedules. 
We report FID and inception score (IS)~\citep{salimans2016improved} scores as metrics. 
As shown in Tab. \ref{tab:cross_schedule}, \methodname achieves significant improvement on linear, quadratic, and cosine schedules both in FID and IS, showing the generality of \methodname in various schedules.

\textbf{Cross-task robustness evaluation.}
We apply \methodname to the text-to-image task for evaluating the generality of our method.
For text-to-image generation, we first introduce 
CLIP~\citep{radford2021learning} to extract the text embedding for MS-COCO~\citep{lin2014microsoft} dataset. Then, DiT-XL/2 is used to train a text-to-image model as our baseline.
Following prior 
work\citep{saharia2022photorealistic}, FID score and CLIP score on MS-COCO validation set are\begin{wraptable}[4]{r}{0.21\textwidth}
\parbox{.21\textwidth}{
    \vspace{-10pt}
    \centering
    
    \tablestyle{3.5pt}{1.3}
     \begin{tabular}{ccc}
         methods  & FID$\downarrow$ & CLIP score$\uparrow$ \\
         \shline 
         \baseline{baseline} & \baseline{27.41} & \baseline{0.237}\\
         \default{\methodname} & \default{\textbf{25.30}} & \default{\textbf{0.244}} \\ 
    \end{tabular}
    \vspace{-5pt}
    \caption{Text to image.}\label{tab:text_to_image}
    }
\end{wraptable} evaluation metrics for quantitative analyses.
As illustrated in Tab.~\ref{tab:text_to_image}, we obtain the better FID and CLIP score than our baseline.%

\subsection{Compatibility with other acceleration methods}
\label{exp:compatibility}
\label{sec:compatibility}

\begin{figure*}[h]
    \centering
    
    \begin{subfigure}{0.42\linewidth}
        \includegraphics[width=1\linewidth]{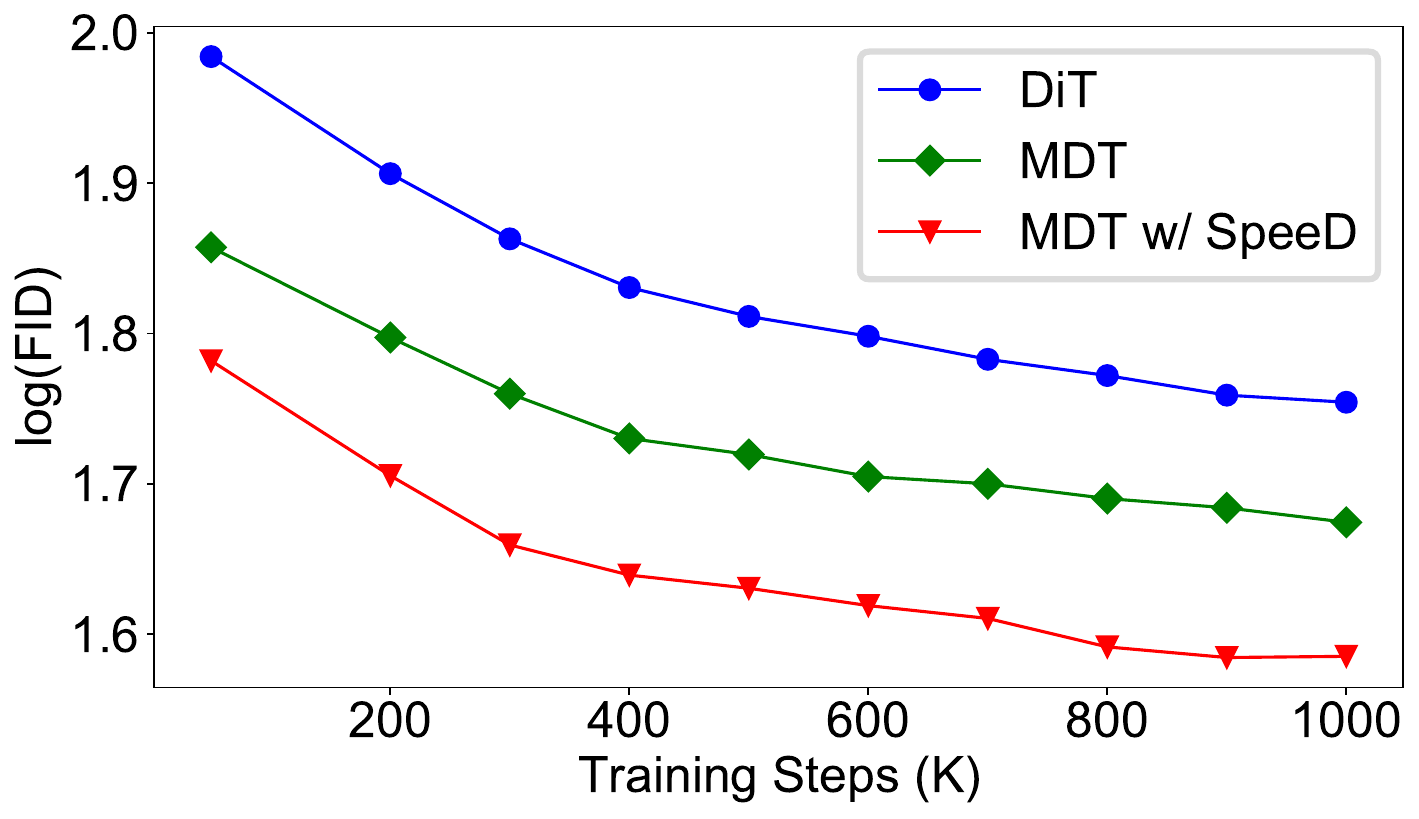}
        \caption{FID-Iteration curve comparisons on \textit{ImageNet-1K}.}
        \label{fig:vs_mdt}
    \end{subfigure}
    \begin{subfigure}{0.42\linewidth}
        \includegraphics[width=1\linewidth]{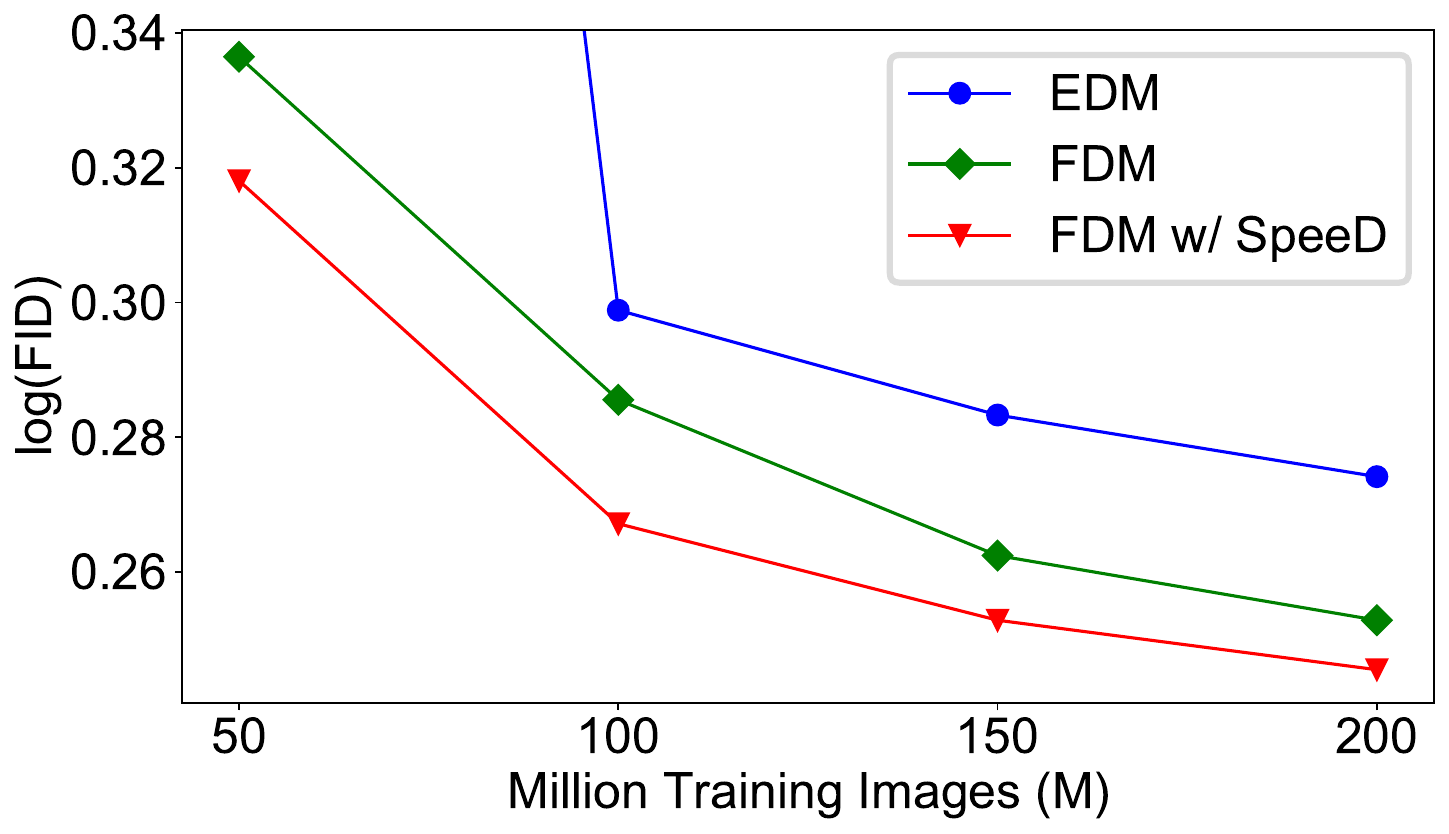}
        \caption{FID-Iteration curve comparisons on \textit{CIFAR-10}.}
        \label{fig:vs_fdm}
    \end{subfigure}
    \vspace{-5pt}

    \caption{SpeeD works well with recent acceleration algorithms and can consistently further accelerate the diffusion model training. We plot the figures using $\log($FID$)$ for better visualization.}
    \label{fig:compatibility_with_others}
    \vspace{-5pt}
\end{figure*}

Until now, we evaluate the effectiveness and generalization of our proposed method: \methodname is a task-agnostic and architecture-agnostic diffusion acceleration approach. 
Is \methodname compatible with other acceleration techniques? To investigate this, we apply our approach with two recent proposed algorithms: masked diffusion transformer (MDT)~\citep{gao2023masked} and fast diffusion model (FDM)~\citep{wu2023fast}.

\textbf{MDT + SpeeD.}
MDT~\citep{gao2023masked} proposes a masked diffusion transformer method, which applies a masking scheme in latent space to enhance the contextual learning ability of diffusion probabilistic models explicitly. MDT can speed up the diffusion training by 10 times.
They evaluate their MDT with DiT-S/2. We just inject our SpeeD on their MDT and report FID-Iteration curves for comparison in Fig.~\ref{fig:vs_mdt}.
All the results are obtained on ImageNet-1K dataset.
Our approach can \textit{further} accelerate MDT at least by \textbf{4} $\times$, which indicates the good compatibility of SpeeD.

\textbf{FDM + SpeeD.}
Fast Diffusion Model~\citep{wu2023fast} is a diffusion process acceleration method inspired by the classic  momentum approach to solve optimization problem in parameter space.
By integrating momentum into the diffusion process, it achieves similar performance as EDM~\citep{karras2022elucidating} with less training cost.
Based on the official implementation, we compare EDM, FDM, and FDM + \methodname on CIFAR-10 of $32 \times 32$ images.
FDM accelerates the EDM by about 1.6 $\times$. \methodname can further reduce the training cost around 1.6 $\times$.

\subsection{Ablation Experiments}

\label{sec:ablation}
We perform extensive ablation studies to illustrate the characteristics of \methodname. The experiments in ablation studies are conducted on the FFHQ dataset and U-Net model by default.  We ablate our designed components: asymmetric sampling (abbreviated as {asymmetric}) and change-aware weighting (abbreviated as {CAW}), suppression intensity $k$ in asymmetric sampling defined in Eqn. \ref{eqn:sampling} and symmetry ceiling $\lambda$ for weighting in Sec.~\ref{ssec:rapid-aware_weighting}.

\begin{table*}[t]
    \begin{subtable}[h]{0.32\textwidth}
    \centering
        \tablestyle{12pt}{1.3}
        \begin{tabular}{ccc}
         sampling curve  & CAW & FID $\downarrow$
         \\
         \shline
         uniform &  & \baseline{17.37} 
         \\ 
         uniform & \checkmark  & 16.75 
         \\ 
         asymmetric &  &  15.82 
         \\  
         asymmetric & \checkmark  & 
         \default{\textbf{15.07}}
         \\
        \end{tabular}
    \caption{\textbf{Components of SpeeD}. Suppressing some trivial time steps does help.}
    \label{tab:abl_component}
    \end{subtable}
    \hfill
    \begin{subtable}[h]{0.32\textwidth}
    \tablestyle{14pt}{1.3}
        \centering
            \begin{tabular}{cc}
         suppression intensity $k$ & FID $\downarrow$
         \\
         \shline
         1 & 15.01 
         \\
          5  & \default{\textbf{14.86}}
         \\
         10 & 16.97 
         \\
         25 & 25.59 
         \\
            \end{tabular}
    \caption{\textbf{Suppression intensity $k$}. Huge suppression decrease diversity to modeling data.}
    \label{tab:abl_k}
    \end{subtable}
    \hfill
    \begin{subtable}[h]{0.32\textwidth}
        \centering
            \tablestyle{18pt}{1.3}
            \begin{tabular}{cc}
        symmetry ceiling $\lambda$ & FID $\downarrow$ 
        \\
         \shline
         0.5 & 15.46
         \\
         0.6  & \default{\textbf{14.86}}
         \\ 
         0.8 & 16.83
         \\
         1.0 & 23.77
         \\
            \end{tabular}
    \caption{\textbf{ Symmetry ceiling $\lambda$}. Weighting served as temperature factor should be in moderation.}
    \label{tab:abl_lambda}
    \end{subtable}
    \vspace{-5pt}
\caption{Ablation studies on FFHQ dataset. Default settings and baseline are in \colorbox{defaultcolor}{purple} and  \colorbox{baselinecolor}{gray}.}
    \vspace{-5pt}
\end{table*}

\paragraph{Evaluating the components of SpeeD.}
Our approach includes two strategies: asymmetric sampling and change-aware weighting. We note these two strategies using `asymmetric' and `CAW'.
We ablate each component in SpeeD. As illustrated in Tab.~\ref{tab:abl_component}, combining our proposed strategies achieves the best results. Using weighting and sampling strategies alone improves the baseline by 0.6 and 1.5 FID scores, respectively, indicating that filtering most samples from the convergence area is more beneficial for training.

\paragraph{Evaluating of suppression intensity $k$.} 
To prioritize the training of critical time steps, asymmetric sampling focus on the time steps out of the convergence area. A higher probability is given for these time steps which is $k$ times than the time steps from the convergence area.
A larger suppression intensity $k$ means a larger gap in training between the different areas of time steps.
We evaluate different suppression intensity $k$ from 1 to 25 and report the FID score In Tab.~\ref{tab:abl_k}.
We observe that k = 5 achieves the best performance. A huge suppression intensity decrease FID scores seriously, which means that it hurts the diversity a lot to modeling data.
This means that the samples in the convergence area, although very close to pure noise, still contains some useful information. Extreme complete discard of these samples results in a degradation of the acceleration performance.

\paragraph{Evaluating of symmetry ceiling $\lambda$.} 
The symmetry ceiling $\lambda$ is a hyper-parameter that regulates the curvature of the weighting function. \textbf{$\lambda$ is set in the interval $[0.5, 1]$}. The midpoint of the re-scaled weight interval is fixed at $0.5$. The symmetry ceiling $\lambda$ is the right boundary of the interval and the left boundary is $1$-$\lambda$. A higher $\lambda$ results in a larger weight interval and a more obvious distinction in weights between different time steps. In Tab. \ref{tab:abl_lambda}, settings $\lambda \leq 0.8 $ obtain higher performance on FID scores than the baseline, which indicates \methodname is relatively robust on symmetry ceiling $\lambda$. Further increase $\lambda$ leads to performance degradation, and Weighting should be in moderation.

\section{Related Work}
We discuss the related works about Diffusion Models and its Training Acceleration. The most related works are as follows, and more discussion are in Appendix \ref{appendix:related}.

\paragraph{Diffusion models} 
Diffusion models have emerged as the dominant approach in generative tasks~\citep{saharia2022photorealistic,chen2023humanmac,wang2024neural,igashov2024equivariant}, which outperform other generative methods including GANs~\citep{zhu2017unpaired,isola2017image,brock2018large}, VAEs~\citep{kingma2013auto}, flow-based models~\citep{dinh2014nice}. These methods~\citep{ho2020denoising, song2020denoising,karras2022elucidating} are based on non-equilibrium thermodynamics~\citep{jarzynski1997equilibrium, sohl2015deep}, where the generative process is modeled as a reverse diffusion process that gradually constructs the sample from a noise distribution~\citep{sohl2015deep}. 
Previous works focused on enhancing diffusion models' generation quality and alignment with users in visual generation. To generate high-resolution images, Latent Diffusion Models (LDMs)~\citep{saharia2022photorealistic,rombach2022high} perform diffusion process in latent space instead of pixel space, which employ VAEs to be encoder and decoder for latent representations. 

\paragraph{Acceleration in diffusion models}
To reduce the computational costs, previous works accelerate the diffusion models in training and inference.
For \textit{training acceleration}, the earliest works~\citep{choi2022perception,hang2023efficient} assign different weights to each time step on Mean Square Error (MSE) loss to improve the learning efficiency. The other methods in training acceleration are proposed, \textit{e.g.}, network architecture~\citep{ryu2022pyramidal,wang2024patch} and diffusion algorithm~\citep{karras2022elucidating,wu2023fast}. Masking modeling~\citep{gao2023masked,zheng2023fast} are recently proposed to train diffusion models. Works
~\citep{go2023addressing,park2024denoising,park2024switch,kim2024denoising} provide observations for explanation from the perspective of multi-tasks learning. 
\methodname is of good compatibility with these methods, \textit{e.g.}, \citep{yu2024unmasking,zheng2024nonuniform, zheng2024beta}.
In the field of sampling acceleration, a number of works tackle the slow inference speed of diffusion models by using fewer reverse steps while maintaining sample quality, including DDIM~\citep{song2020denoising}, Analytic-DPM~\citep{bao2022analytic}, and DPM-Solver~\citep{lu2022dpm}.

\paragraph{Conditional generation.} To better control the generation process with various conditions, \textit{e.g.}, image style, text prompt and stroke, Classifier-Free Guidance (CFG) proposes a guidance strategy with diffusion models that balance the sample quality and prompt alignment. ControlNet~\citep{zhang2023adding} reuses the large-scale pre-trained layers of source
models to build a deep and strong encoder to learn specific conditions.
Recently benefiting from diffusion models in the image generation field, video generation is getting trendy. The promising results are provided in recent works~\citep{ma2024latte,pku_yuan_lab_and_tuzhan_ai_etc_2024_10948109} on video generation tasks.

\section{Conclusion}

\methodname, a approach for accelerating diffusion training by closely examining time steps is proposed. The core insights are: 1) suppressing the sampling probabilities of time steps that offer limited benefits to diffusion training (i.e., those with extremely small losses), and 2) emphasizing the importance of time steps with rapidly changing process increments.
\methodname demonstrates strong robustness across various architectures and datasets, achieving significant acceleration on multiple diffusion-based image generation tasks.
Extensive theoretical analyses in this paper, are provided.

Referring to previous works~\citep{peebles2023scalable,go2023addressing,park2024denoising,choi2022perception,hang2023efficient,esser2024scaling,xu2024towards}, our experiments train 50K, 100K, and 400K in MetFaces 256 × 256, FFHQ 256 × 256, and ImageNet-1K, respectively. 7M iterations are used in DiT to train ImageNet in the original work~\cite{peebles2023scalable}, which 1) is not the amount of resource usage that can be achieved in general research. Meanwhile, 2) the focus is on the acceleration effect of training, and a direct comparison of the ultimate convergence stage is unnecessary for this work.

\paragraph{Acknowledgments.} This research is supported by the National Research Foundation, Singapore under its AI Singapore Programme (AISG Award No: AISG2-PhD-2021-08-008). This work is partially supported by the National Natural Science Foundation of China (62176165), the Stable Support Projects for Shenzhen Higher Education Institutions (20220718110918001), the Natural Science Foundation of Top Talent of SZTU(GDRC202131). Yang You's research group is being sponsored by NUS startup grant (Presidential Young Professorship), Singapore MOE Tier-1 grant, ByteDance grant, ARCTIC grant, SMI grant (WBS number: A-8001104-00-00),  Alibaba grant, and Google grant for TPU usage. We thank Tianyi Li, Yuchen Zhang, Yuxin Li, Zhaoyang Zeng, and Yanqing Liu for the comments on this work. Xiaojiang Peng, Hanwang Zhang, and Yang You are equal advising. Xiaojiang Peng is the corresponding author.
{
    \small
    \bibliographystyle{sec_CVPR2025/ieeenat_fullname}
    \bibliography{references}
}

\newpage
\appendix
\clearpage
\setcounter{page}{1}
\maketitlesupplementary

\section{More Detail of Experiments}
\label{details_of_training_appendix}
In this section, we introduce detailed experiment settings, datasets, and architectures.

\nocite{zheng2023enfomax}
\nocite{zheng2023fast}
\nocite{zheng2023mcae}
\nocite{zheng2024beta}
\nocite{zheng2024famim}
\nocite{zheng2024non}
\nocite{zheng2024nonuniform}

\subsection{Architecture and Training Recipe.}
\label{arch_details}
We utilize Unet and DiT as our base architecture in the diffusion model. pre-trained VAE which loads checkpoints from \href{https://huggingface.co/stabilityai/sd-vae-ft-mse}{huggingface} is employed to be latent encoder. Following Unet implementation from \href{https://github.com/CompVis/stable-diffusion}{LDM} and DIT from official implementation, we provide the architecture detail in Tab. \ref{tab:arch_conf}. We provide our basic training recipe and evaluation setting with specific details in Tab. \ref{tab:detail_conf}. 

\begin{table*}[htp]
    \centering
     \tablestyle{15pt}{1.3}
     \begin{tabular}{ccccccc}
    architecture & input size & input channels & patch size & model depth & hidden size &  attention heads \\
    \shline
    U-Net & 32 $\times$ 32 & 3 & - & 8 & 128 & 1 \\
    DiT-XL/2 & 32 $\times$ 32 & 4 & 2 & 28 & 1152 & 16 \\ 
    DiT-S/2 & 32 $\times$ 32 & 4 & 2 & 12 & 384 & 6 \\
    \end{tabular}
    \caption{Architecture detail of Unet and DiT on MetFaces, FFHQ, and ImageNet.}
    \label{tab:arch_conf}
\end{table*}

\begin{table*}[htp]
    \vspace{-10pt}
    \centering
     \tablestyle{22pt}{1.3}
     \begin{tabular}{cccc}
     & MetFaces 256 $\times$ 256 & FFHQ 256 $\times$ 256  & ImageNet-1K 256 $\times$ 256 \\
    \shline
     latent size & 32 $\times$ 32 $\times$ 3 & 32 $\times$ 32 $\times$ 3 & 32 $\times$ 32 $\times$ 3 \\
     class-conditional & & & \checkmark \\
     diffusion steps & 1000 & 1000 & 1000 \\
     noise schedule & linear & linear & linear \\
     batch size & 256 & 256 & 256 \\
     training iterations & 50K & 100K & 400K \\
     optimizer & AdamW & AadamW & AdamW \\
     learning rate & 1e-4 & 1e-4 & 1e-4 \\
     weight decay & 0 & 0 & 0 \\ 
     sample algorithm & DDPM & DDPM & DDPM \\
     number steps in sample & 250 & 250 & 250 \\
     number sample in evaluation & $10,000$ & $10,000$ & $10,000$ \\

    \end{tabular}
    \caption{Our basic training recipe based on MetFaces, FFHQ, ImageNet datasets}
    \label{tab:detail_conf}
\end{table*}

\subsection{Datasets}
\textbf{CIFAR-10}. CIFAR-10 datasets consist of 32 $\times$ 32 size colored natural images divided into categories. It uses $50,000$ in images for training and \href{https://github.com/NVlabs/edm}{EDM evaluation suite} in image generation.

\textbf{MetFaces} is an image dataset of human faces extracted from works of art. It consists of 1336 high-quality PNG images at 1024×1024 resolution. We download it at 256 resolution from \href{https://www.kaggle.com/}{kaggle}.

\textbf{FFHQ} is a high-quality image dataset of human faces, contains 70,000 images. We download it at 256x256 resolution from \href{https://www.kaggle.com/}{kaggle}.

\textbf{ImageNet-1K} is the subset of the ImageNet-21K dataset with $1,000$ categories. It contains $1,281,167$ training images and $50,000$ validation images. 

\textbf{MSCOCO} is a large-scale text-image pair dataset. It contains 118K training text-image pairs and 5K validation images. We download it from \href{https://cocodataset.org/#home}{official website}.

\textbf{FaceForensics} is a video dataset consisting of more than 500,000 frames containing faces from 1004 videos that can be used to study image or video forgeries.

\subsection{Detail of MDT + SpeeD Experiment}
MDT utilizes an asymmetric diffusion transformer architecture, which is composed of three main components: an encoder, a side interpolater, and a decoder. During training, a subset of the latent embedding patches is randomly masked using Gaussian noise with a masking ratio. Then, the remaining latent embedding, along with the full latent embedding is input into the diffusion model. 

Following the official implementation of MDT, We utilize DiT-S/2 and MDT-S/2 as our base architecture, whose total block number both is 12 and the number of decoder layers in MDT is $2$. We employ the AdamW~\citep{loshchilov2017decoupled} optimizer with constant learning rate 1e-4 using $256$ batch size without weight decay on class-conditional ImageNet with an image resolution of $256^2$. We perform training on the class-conditional ImageNet dataset with images of resolution 256x256. The diffusion models are trained for a total of 1000K iterations, utilizing a mask ratio of 0.3. 

\begin{table*}[t]
    \centering
    \tablestyle{18pt}{1.3}
    \caption{The ingredients of generalized curves $\dot{\Delta}$ and $\dot{\Sigma}$ schedules about mainstream SDE designs, including VP, VE~\citep{song2021score}, EDM~\citep{karras2022elucidating}.}
    \begin{tabular}{ccccc}
        Schedules
        &
        $s$
        &
        $\sigma^{2}$
        &
        $\dot{s}$
        &
        $\dot{\sigma}$
        \\ \shline
        VP
        & 
        $\exp\{-\frac{1}{4}\Delta_{\beta}t^{2}-\frac{1}{2}\beta_{0}t\}$
        &
        $\exp\{\frac{1}{2}\Delta_{\beta}t^{2}+\beta_{0}t\}-1$
        &
        $-\frac{\sigma\dot{\sigma}}{(1+\sigma^{2})^{3/2}}$
        &
        $\frac{(1+\sigma^{2})(\Delta_{\beta}t+\beta_{0})}{2\sigma}$
        \\
        VE
        &
        $1$
        &
        $t$
        &
        $0$
        &
        $1$
        \\
        EDM
        &
        $1$
        &
        $t^2$
        &
        $0$
        &
        $2 t$
    \end{tabular}
    \vspace{-10pt}
    \label{tab:ingredients}
\end{table*}
\subsection{Detail of FDM + SpeeD Experiment}
FDM add the momentum to the forward diffusion process with a scale that control the weight of momentum for faster convergence to the target distribution. 
Following official implementation, we train diffusion models of EDM and FDM. We retrain these official network architecture which is U-Net with positional time embedding with dropout rate 0.13 in training. We adopt Adam optimizer with learning rate 1e-3 and batch size 512 to train each model by a total of 200 million images of $32^2$ CIFAR-10 dataset. During training, we adopt a learning rate ramp-up duration of 10 Mimgs and set the EMA half-life as 0.5 Mimgs. For evaluation, EMA models generate 50K images using EDM sampler based on Heun's $2^{nd}$ order method~\citep{suli2003introduction}. 

\subsection{Text-to-Image Experiment Detail}
In text to image task, diffusion models synthesize images with textual prompts. For understanding textual prompts, text-to-image models need semantic text encoders to encode language text tokens into text embedding. We incorporate a pre-trained CLIP language encoder, which processes text with a maximum token length of 77. DiT-XL/2 is employed as our base diffusion architecture. We employ AdamW optimizer with a constant learning rate 1e-4 without weight decay. We train text-to-image diffusion models for 400K training iterations on MS-COCO training dataset and evaluate the FID and CLIP score on MS-COCO validation dataset. To enhance the quality of conditional image synthesis, we implement classifier-free guidance with 1.5 scale factor.

\newtheorem{proposition}{Proposition}[section]
\subsection{Theoretical Analysis}

\subsubsection{Notations}
\label{appdx:sssec_notations}

In this section, we will introduce the main auxiliary notations and the quantities that need to be used. The range of schedule hyper-parameter group $\{\beta_{t}\}_{t\in[T]}$ turns out to be $t=1$ to $t=T$. For analytical convenience, we define $\beta_{0}$ as
$\beta_{0}:=\beta_{1}-\Delta_{\beta}/T$.

Another auxiliary notation is forward ratio $\rho_{t}$, which is defined as $\rho_{t}=t/T$. Forward ratio provide an total number free notation for general diffusion process descriptions.

Based on the two auxiliary notations $\beta_{0}$ and $\rho_{t}$, the expression of $\beta_{t}$ with respect to the forward process ratio is
$\beta_{t}=\beta_{0}+\Delta_{\beta}\rho_{t}$.

The relationship between $\alpha_{t}$ and $\beta_{t}$ is recalled and re-written as follows:
$\alpha_{t}=1-\beta_{t}=1-\beta_{0}-\Delta_{\beta}\rho_{t}$.
$\bar{\alpha}_{t}$ the multiplication of $\alpha_{t}$ is re-written as $\bar{\alpha}_{t}=\Pi_{s=1}^{t}(1-\beta_{0}-\Delta_{\beta}\rho_{s})$.

Perturbed samples' distribution:
$x_{t}|x_{0}\sim\mathcal{N}(\sqrt{\bar{\alpha}_{t}}x_{0},(1-\bar{\alpha}_{t})\mathbf{I})$

\subsubsection{Auxiliary Lemma and Core Theorem}

\begin{lemma}[Bounded $\alpha$ by $\beta$]
    \label{lemma:a_b}
    In DDPM~\citep{ho2020denoising}, using a simple equivariant series $\{\beta_{t}\}_{t\in [T]}$ to simplify the complex cumulative products $\{\bar{\alpha}_{t}\}_{t\in [T]}$, we obtain the following auxiliary upper bound of $\bar{alpha}_{t}$.
    \begin{align*}
    \bar{\alpha}_{t}
    &
    \leq
    \exp\{-(\beta_{0}t+\frac{\Delta_{\beta}t^{2}}{2T})\}
    \end{align*}
\end{lemma}

\subsubsection{Propositions}

\begin{proposition}[Jensen's inequality]
\label{appdx_prop_jensen}
If $f$ is convex, we have:
$$
\begin{aligned}
    \mathbf{E}_{X}f(X) \ge f(\mathbf{E}_{X}X).
\end{aligned}
$$
A variant of the general one shown above:
$$
\begin{aligned}
    ||\sum_{i\in[N]}x_{i}||^{2} \le 
 N\sum_{i\in[N]}||x_{i}||^{2}.
\end{aligned}
$$
\end{proposition}
\begin{proposition}[triangle inequality]
\label{appdx_prop_tri}
    The triangle inequality is shown as follows, where $||\cdot||$ is a norm and $A,B$ is the quantity in the corresponding norm space:
    $$||A+B||\le ||A||+||B||$$.
\end{proposition}

\begin{proposition}[matrix norm compatibility]
\label{appdx_prop_mnc}
     The matrix norm compatibility, $A\in \mathbb{R}^{a\times b}, B\in \mathbb{R}^{b\times c}, v\in \mathbb{R}^{b}$:
     $$
     \begin{aligned}
        ||AB||_{m}\le ||A||_{m}||B||_{m} \\
        ||Av||_{m}\le ||A||_{m}||v||.
     \end{aligned}
     $$
\end{proposition}

\begin{proposition}[Peter Paul inequality]
\label{appdx_prop_ppi}
    $$ 2 \langle x, y \rangle \le \frac{1}{\epsilon}||x||^{2} + \epsilon ||y||^{2}$$.
\end{proposition}

\subsubsection{Proof of Lemma~\ref{lemma:a_b}}
\begin{proof}
To proof the auxiliary Lemma~\ref{lemma:a_b}, we re-arrange the notation of $\bar\alpha_{t}$ as shown in Section~\ref{appdx:sssec_notations}, and we have the following upper bound:
\begin{align*}
    \log\bar{\alpha}_{t}
    &=
    \sum_{s=1}^{t}\log(1-\beta_{0}-\Delta_{\beta}\rho_{s})
    \\
    &\leq
    t\log(\frac{1}{t}\sum_{s=1}^{t}(1-\beta_{0}-\Delta_{\beta}\rho_{s}))
    \\
    &=
    t\log(1-\beta_{0}-\Delta_{\beta}\frac{1}{t}\sum_{s=1}^{t}\frac{s}{T})
    \\
    &=
    t\log(1-\beta_{0}-\Delta_{\beta}\frac{t+1}{2T})
    \\
    &\leq
    -(\beta_{0}t+\frac{\Delta_{\beta}(t+1)t}{2T}),
\end{align*}
where the two inequalities are by the concavity of $\log$ function and the inequality: $\log(1+x)\leq x$. Taking exponents on both sides simultaneously, we have:
\begin{align*}
    \bar{\alpha}_{t}
    &
    \leq
    \exp\{-(\beta_{0}t+\frac{\Delta_{\beta}t^{2}}{2T})\}.
\end{align*}
\end{proof}

\subsubsection{Proof of Theorem~\ref{theo:bound}}

Before the proof of the theorem, we note that the samples $x_{t}|x_{0}\sim \mathcal{N}(\mu_{t},\mathbf{\sigma}_{t})$ have the following bounds with Lemma~\ref{lemma:a_b}:
\begin{itemize}
    \item 
    Reformulate the expression of $\sqrt{\bar{\alpha}}$, we have the mean vector $\mu_{t}$'s components $\dot{\mu}_{t}$ bounded by $\dot{x}_{0}$ the corresponding components of data $x_{0}$ as follows:
    
    $$\dot{\mu}_{t}=\sqrt{\bar{\alpha}_{t}}\dot{x}_{0}
    \leq
    \exp\{-\frac{1}{2}(\beta_{0}t+\frac{\Delta_{\beta}t^{2}}{2T})\}\dot{x}_{0},$$
    
    \item Reformulate the expression of $\bar{\alpha}$, we have a partial order relation on the cone about covariance matrix of $x_{t}|x_{0}$ as follows:
$$\mathbf{\sigma}_{t}=(1-\bar{\alpha}_{t})\mathbf{I}\succeq (1-\exp\{-(\beta_{0}t+\frac{\Delta_{\beta}t^{2}}{2T})\})\mathbf{I}.$$
\end{itemize}

\begin{proof}
The process increment at given $t^{\text{th}}$ time step is $\delta_{t}=x_{t+1} - x_{t}$. $\delta_{t}$ is a Gaussian process as follows:
$$\delta_{t}\sim\mathcal{N}(\underbrace{(\sqrt{\alpha_{t+1}}-1)\sqrt{\bar{\alpha}_{t}}x_{0}}_{\phi_{t}},\underbrace{[2-\bar{\alpha}_{t}(1+\alpha_{t+1})]\mathbf{I}}_{\Psi_{t}})$$

The theorem's key motivation is that the label is noisy, and noisy magnitude is measured by mean vector's norm $||\phi_{t}||$ and covariance matrix $\Psi_{t}$.

The upper bounds of mean vectors' norm and the partial order of covariance matrix at different time step $t$ are shown as follows: 
\begin{align*}
||\phi_{t}||^{2}
&
\leq
(\sqrt{\alpha_{t+1}}-1)^{2}\bar{\alpha}_{t}||\mathbb{E}x_{0}||^{2}
\\
&
\leq
(1-\alpha_{t+1})\bar{\alpha}_{t}||\mathbb{E}x_{0}||^{2}
\\
&
\leq
(\underbrace{\beta_{0}+\Delta_{\beta}\rho_{t+1}}_{\beta_{t+1}})\exp\{-(\underbrace{\beta_{0}+\frac{\Delta_{\beta}t}{2T}}_{\beta_{t/2}})t\}||\mathbb{E}x_{0}||^{2}
\\
&
\leq
\beta_{\max}\exp\{-(\underbrace{\beta_{0}+\frac{\Delta_{\beta}t}{2T}}_{\beta_{t/2}})t\}||\mathbb{E}x_{0}||^{2}
\end{align*}
where the inequalities are by Lemma~\ref{lemma:a_b}, $(1-x)^{2}\leq(1-x^{2})=(1-x)(1+x)$, when $x\in[0,1]$, and $\beta_{t+1}\leq \beta_{\max}$ 

\begin{align*}
\Psi_{t}
&
=
[2(1-\bar{\alpha}_{t})+\bar{\alpha}_{t}(\beta_{0}+\Delta_{\beta}\rho_{t+1})]\mathbf{I}
\\
&
\succeq
2(1-\exp\{-(\underbrace{\beta_{0}+\frac{\Delta_{\beta}t}{2T}}_{\beta_{t/2}})t\})\mathbf{I}
+
\bar{\alpha}_{t}\beta_{t+1}
\mathbf{I}
\\
&
\succeq
2(1-\exp\{-(\underbrace{\beta_{0}+\frac{\Delta_{\beta}t}{2T}}_{\beta_{t/2}})t\})\mathbf{I}
\end{align*}
where the inequalities are by Lemma~\ref{lemma:a_b} and $\bar{\alpha}_{t}\beta_{t+1}\mathbf{I}\succeq \mathbf{0}$. The residual term is  $$\bar{\alpha}_{t}\beta_{t+1}=\beta_{t+1}\Pi_{s=1}^{t}(1-\beta_{s})\geq\exp\{\log\beta_{t+1}+t\log(1-\beta_{t})\}$$
\end{proof}

\section{More Experiment Results}
\label{more_results_appendix}

\begin{figure*}[h]
    
    \centering
    \begin{subfigure}{0.19\linewidth}
        \includegraphics[width=1\linewidth]{figures/Metfaces_dit.pdf}
    \end{subfigure}
    \begin{subfigure}{0.19\linewidth}
        \includegraphics[width=1\linewidth]{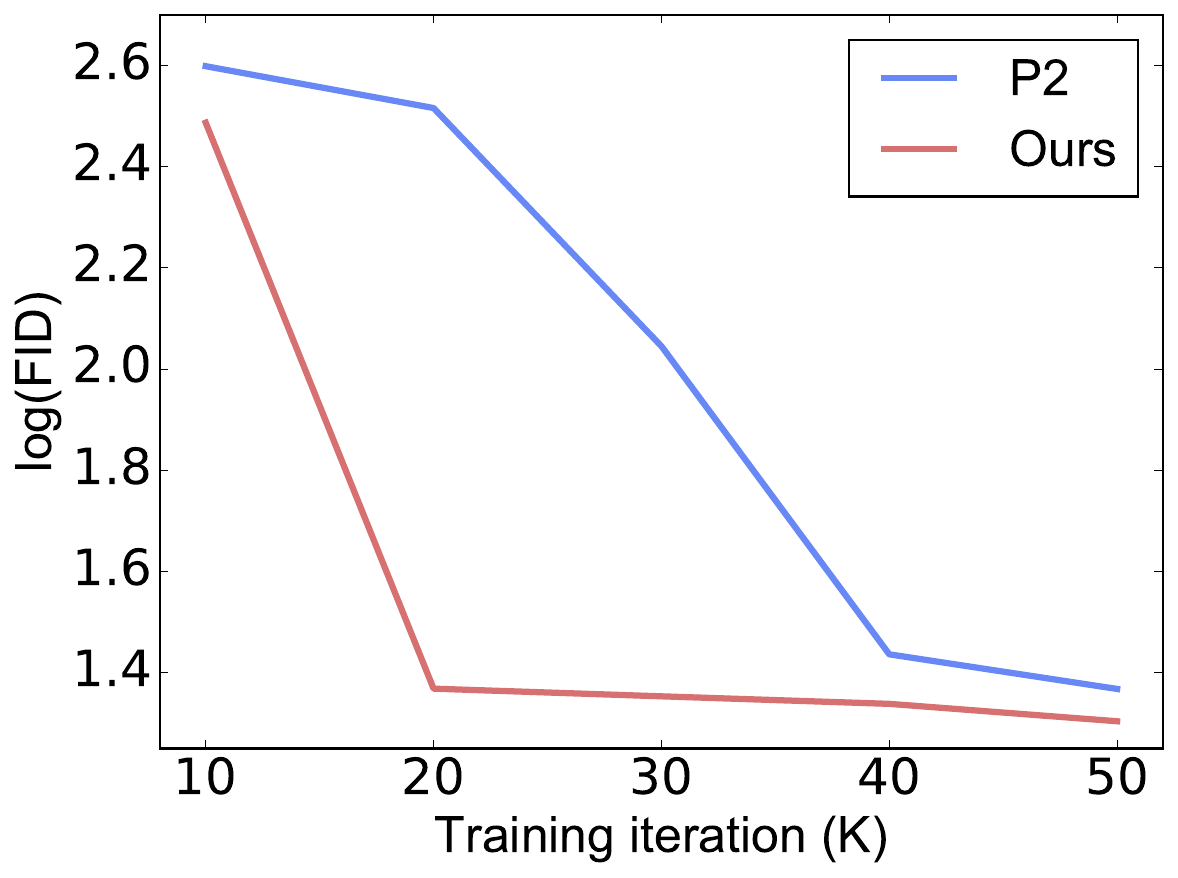}
    \end{subfigure}
     \begin{subfigure}{0.19\linewidth}
        \includegraphics[width=1\linewidth]{figures/Metfaces_Min-SNR.pdf}
    \end{subfigure}
     \begin{subfigure}{0.19\linewidth}
        \includegraphics[width=1\linewidth]{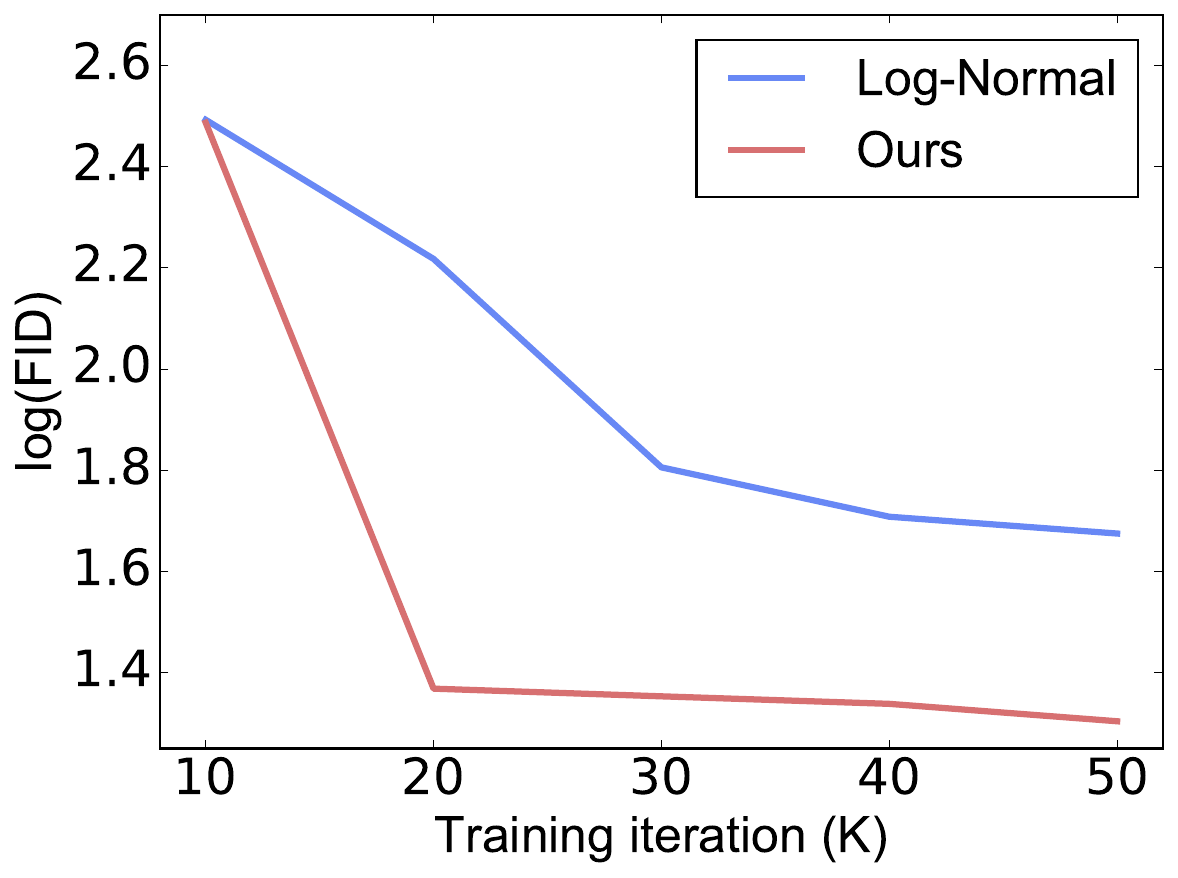}
    \end{subfigure}
     \begin{subfigure}{0.19\linewidth}
        \includegraphics[width=1\linewidth]{figures/Metfaces_CLTS.pdf}
    \end{subfigure}

    \caption{More efficiency comparison on MetFaces.}

    \label{fig:efficiency_appendix}
    
\end{figure*}

\paragraph{Efficiency comparisons.}
In Fig.~\ref{fig:efficiency_appendix}, besides the Min-SNR and CLTS, we show the efficiency comparison with P2 and Log-Normal methods.
One can find that our method consistently accelerates the diffusion training in large margins.

\begin{wraptable}[5]{r}{0.21\textwidth}
\parbox{.21\textwidth}{
    \vspace{-10pt}
    \centering
    \caption{Super resolution.}
    \vspace{-8pt}
    \tablestyle{6pt}{1.3}
     \begin{tabular}{ccc}
         Method  & 50K & 100K \\
         \shline 
         \baseline{DiT-XL/2} & \baseline{77.9} & \baseline{35.4} \\
         \default{SpeeD} & \default{48.7} & \default{10.6} \\ 
    \end{tabular}
    \label{tab:super_resolution}
    }
\end{wraptable}

\textbf{Super resolution with SpeeD.} We employ SpeeD to super-resolution image generation on 512 $\times$ 512 MetFaces compared with vanilla DiT. We train DiT-XL/2 for 100K training iterations and compare the FID score at 50K, 100K training iterations. The batch size is $32$ for saving the GPU memory. As shown in \ref{tab:super_resolution}, SpeeD obtain better performance than vanilla DiT at same training iterations on $512^2$ MetFaces dataset. It indicates that SpeeD can achieve training acceleration on super-resolution tasks.

\subsection{Additional Experiments}
\label{appdx_add_exp}
\paragraph{Experimetns on DiT-S/8.} Further comparison on DiT of smaller scales are reported in Tab.~\ref{tab:tbl_ImageNet_Celeb-A}. The datasets include ImageNet-1K and Celeb-A.

\begin{table}[h]
\tablestyle{5pt}{1.1}
    \centering
    \begin{tabular}{ccccccc}
        \multicolumn{6}{c}{ImageNet-1K (FID $\downarrow$)}
        \\
        \#steps
        &
        \baseline{DiT-S/8}
        &
        P2
        &
        Min-SNR
        &
        Log-normal
        &
        CLTS
        &
        \default{ours}
        \\
        10K
        &
        \baseline{399.9}
        &
        398.6
        &
        398.4
        &
        399.7
        &
        399.6
        &
        \default{400.8}
        \\
        20K
        &
        \baseline{380.0}
        &
        365.0
        &
        368.0
        &
        387.5
        &
        381.9
        &
        \default{379.2}
        \\
        40K
        &
        \baseline{\underline{200.0}}
        &
        207.6
        &
        208.6
        &
        365.9
        &
        231.6
        &
        \default{\textbf{191.5}}
        \\
        \shline
        \multicolumn{6}{c}{Celeb-A (FID $\downarrow$)}
        
        \\
        10K
        &
        \baseline{408.7}
        &
        412.5
        &
        408.7
        &
        410.4
        &
        408.0
        &
        \default{407.8}
        \\
        20K
        &
        \baseline{386.7}
        &
        366.8
        &
        386.7
        &
        394.0
        &
        386.4
        &
        \default{377.9}
        \\
        40K
        &
        \baseline{271.6}
        &
        271.1
        &
        271.6
        &
        293.0
        &
        \underline{258.8}
        &
        \default{\textbf{254.9}}
        \\
        \shline
        \multicolumn{6}{c}{Celeb-A (IS $\uparrow$)}
        
        \\
        10K
        &
        \baseline{1.50}
        &
        1.49
        &
        1.50
        &
        1.50
        &
        1.50
        &
        \default{1.50}
        \\
        20K
        &
        \baseline{1.49}
        &
        1.48
        &
        1.49
        &
        1.50
        &
        1.49
        &
        \default{1.63}
        \\
        40K
        &
        \baseline{3.29}
        &
        \underline{3.70}
        &
        3.29
        &
        2.55
        &
        3.46
        &
        \default{\textbf{3.87}}
    \end{tabular}
    \vspace{-10pt}
    \caption{Comparison on ImageNet-1K and Celeb-A. FID and IS are reported. The baseline is measured on DiT-S/8, with global batchsize of 16. Other settings are default.}
    \vspace{-15pt}
    \label{tab:tbl_ImageNet_Celeb-A}
\end{table}

\paragraph{More baslines.} Comparison between SpeeD and BS~\citep{zheng2024nonuniform} and B-TTDM~\citep{zheng2024beta} are shown in Tab.~\ref{tab:additional_baselines}.

\paragraph{Detailed ablation study.} FID-10K on 10K/20K/40K/50K iterations are provided in Tab.~\ref{tab:ablation_iterations} the model is DiT-S/8 with batchsize of 16.

\begin{table}[t]
\tablestyle{14pt}{1.1}
    \centering
    \begin{tabular}{cccc}
        \#steps
        &
        BS~\citep{zheng2024nonuniform}
        &
        B-TTDM~\citep{zheng2024beta}
        &
        \default{ours}
        \\
        \shline
        100K
        & 
        155.9
        & 
        157.4
        & 
        \default{\textbf{155.2}}
        \\
        200K
        &
        152.2
        &
        \textbf{150.9}
        &
        \default{\underline{151.7}}
        \\
        400K
        &
        141.8
        &
        140.5
        &
        \default{\textbf{139.2}}
    \end{tabular}
    \vspace{-10pt}
    \caption{New baselines to be added. The settings follow BS. (FID)}
    \label{tab:additional_baselines}
\end{table}
\begin{table}[t]
\tablestyle{12pt}{1.1}
\vspace{-10pt}
    \centering
    \begin{tabular}{ccccc}
         \#steps
         &
         10K
         &
         20K
         &
         40K
         &
         50K
         \\
         \baseline{DiT-S/8}
         &
         \baseline{399.9}
         &
         \baseline{380.0}
         &
         \baseline{200.0}
         &
         \baseline{--}
         \\
         \shline
         $\lambda = 0.5$
         &
         400.7
         &         
         376.7
         &
         207.4
         &
         202.1
         \\
         \default{$\lambda = 0.6$}
         &
         \default{400.8}
         &         
         \default{379.2}
         &
         \default{191.5}
         &
         \default{191.2}
         \\
         $\lambda = 0.8$
         &
         400.3
         &         
         379.2
         &
         203.2
         &
         200.5
         \\
         \shline
         $\tau=600$
         &
         401.0
         &         
         382.6
         &
         210.1
         &
         198.5
         \\
         \default{$\tau=700$}
         &
         \default{400.8}
         &         
         \default{379.2}
         &
         \default{191.5}
         &
         \default{191.2}
         \\
         $\tau=800$
         &
         399.5
         &         
         380.9
         &
         200.1
         &
         200.0
         \\
         \shline
         $k=1$
         &
         400.4
         &         
         388.0
         &
         214.5
         &
         202.4
         \\
         \default{$k=2$}
         &
         \default{400.8}
         &         
         \default{379.2}
         &
         \default{191.5}
         &
         \default{191.2}
         \\
         $k=10$
         &
         400.4
         &         
         380.5
         &
         231.7
         &
         206.7
    \end{tabular}
    \vspace{-10pt}
    \caption{Detailed ablation across training steps. FID-10K is reported. DiT-S/8 serves as the baseline. Global batchsize is 16.}
    \vspace{-15pt}
    \label{tab:ablation_iterations}
\end{table}

\subsubsection{Detailed Training Process}
The detailed training process on FFHQ through 100K iterations are shown in Tab.~\ref{tab:detailed_process}.
\begin{table*}[t]
    \centering
    \begin{tabular}{c|cccccccccc}
iterations (K) &	10 &	20 &	30 &	40 &	50 &	60 &	70 &	80 &	90 &	100 \\
\shline
DiT-XL/2 &	356.1 &	335.3 &	165.2 &	35.8 &	12.9 &	11.9 &	10.5 &	9.6 &	8.7 &	7.8 \\
SpeeD &	322.1 &	320.0 &	91.8 &	19.8 &	9.9 &	7.6 &	7.1 &	6.6 &	6.2 &	5.8 \\
    \end{tabular}
    \caption{Details about training to 100K on FFHQ.}
    \label{tab:detailed_process}
\end{table*}

\section{More Related Works}
\label{appendix:related}
We discuss other works related to SpeeD, including Text to Image and Video generation. Another point to mention is that we learn from InfoBatch~\citep{qin2023infobatch} in writing.

\textbf{Text to image generation with diffusion models} Text-to-image generation has emerged as a hotly contested and rapidly evolving field in recent years, with an explosion of related industrial products springing up~\citep{saharia2022photorealistic,rombach2022high,betker2023improving,chen2023pixart,esser2024scaling}. Convert textual descriptions into corresponding visual content, models not only learn to synthesize image content but also ensuring alignment with the accompanying textual descriptions. To better align images with textual prompt guidance, previous work has primarily focused on enhancements in several schemes including strengthening the capacity of text encoder~\citep{raffel2020exploring,radford2021learning}  improving the condition plugin module in diffusion model~\citep{zhang2023adding}, improving data quality~\citep{betker2023improving}. 

\textbf{Video generation with diffusion models.}
As diffusion models achieve tremendous success in image generation, video generation has also experienced significant breakthroughs, marking the field's evolution and growth. 
Inspired by image diffusion, pioneering works such as RVD~\cite{RVD} and VDM~\cite{VDM} explore video generation using diffusion methods. Utilizing temporal attention and latent modeling mechanisms, video diffusion has advanced in terms of generation quality, controllability, and efficiency~\cite{ho2022imagen, MakeAVideo, Zhou2022MagicVideoEV, LVDM, show1, Animatediff, TuneAVideo, videocomposer2023}. Notably, Stable Video Diffusion~\cite{SVD} and Sora~\cite{videoworldsimulators2024} achieve some of the most appealing results in the field.

\textbf{Other diffusion acceleration works}
To achieve better results with fewer NFE steps, Consistency Models~\cite{song2023consistency} and Consistency Trajectory Models~\cite{kim2023consistency} employ consistency loss and novel training methods. Rectified Flow~\cite{liu2022flow}, followed by Instaflow~\cite{liu2023instaflow}, introduces a new perspective to obtain straight ODE paths with enhanced noise schedule and improved prediction targets, together with the reflow operation. DyDiT~\cite{zhao2024dynamic} incorporates dynamic neural networks~\cite{han2021dynamic, zhao2024dynamic, zhao2024stitch} into diffusion models, achieving significant acceleration.

\section{Visualization}
\label{vis_appendix}

\begin{figure*}[h]
    \centering
    \includegraphics[width=\textwidth]{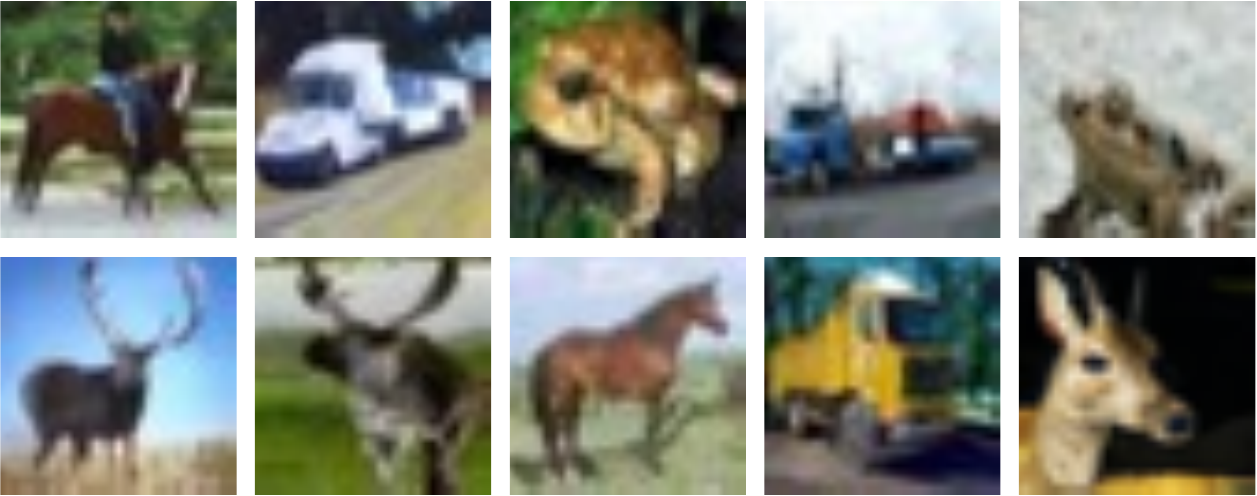}
    \caption{Generated images of CIFAR-10.}
    \label{fig:vis_cifar}
\end{figure*}

\begin{figure*}[h]
    \centering
    \includegraphics[width=\textwidth]{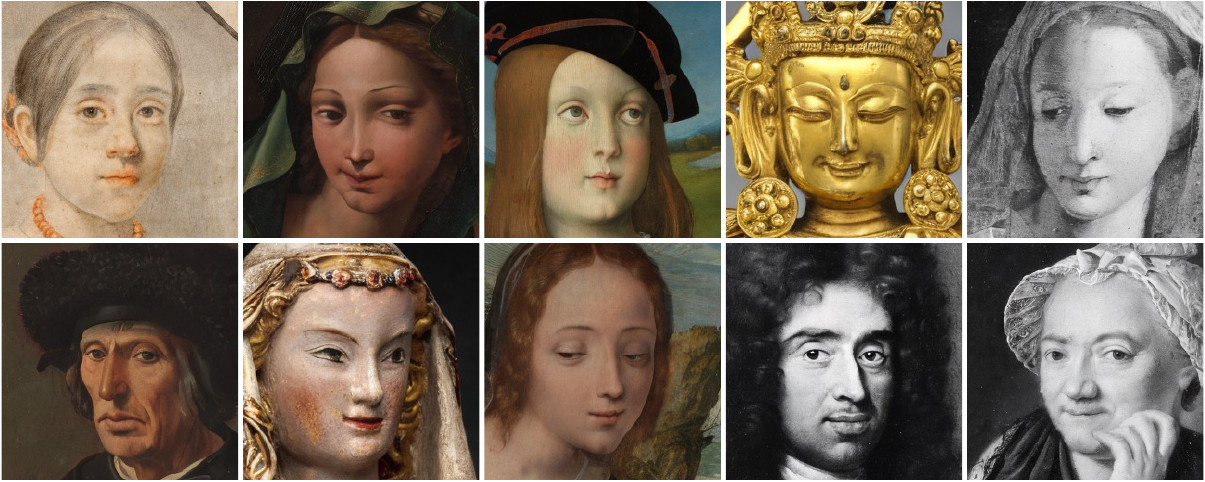}
    \caption{Generated images of MetFaces.}
    \label{fig:vis_metface_appendix}
\end{figure*}

\begin{figure*}[h]
    \centering
    \includegraphics[width=\textwidth]{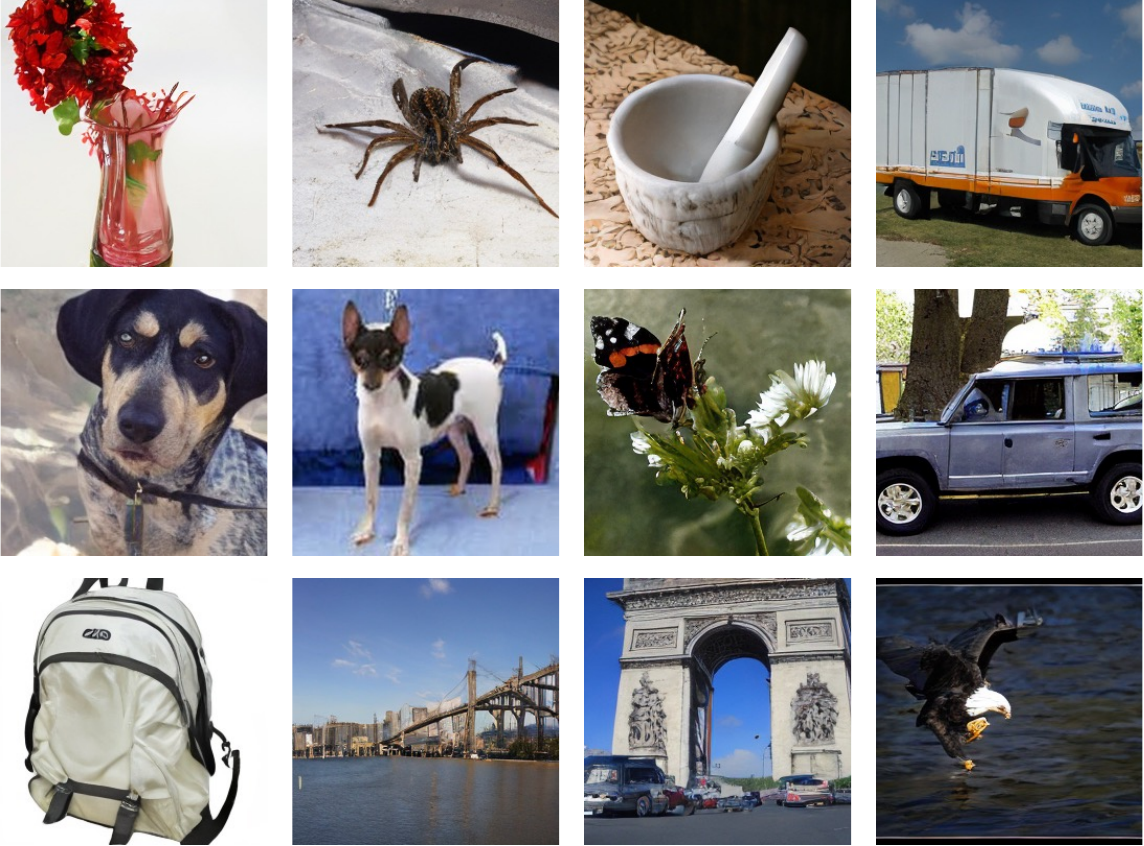}
    \caption{Generated images of ImageNet-1K.}
    \label{fig:vis_imagenet}
\end{figure*}

\paragraph{Visualizations of the generated images.}
The figures above illustrate the quality of images generated by our method across various datasets, including CIFAR-10, FFHQ, MetFaces, and ImageNet-1K. In Fig.~\ref{fig:vis_cifar}, the generated images from the CIFAR-10 dataset display distinct and recognizable objects, even for challenging categories. Fig.~\ref{fig:vis_ffhq} presents generated images from the FFHQ dataset, showcasing diverse and realistic human faces with varying expressions and features. Fig.~\ref{fig:vis_metface_appendix} exhibits images from the MetFaces dataset, depicting detailed and lifelike representations of artistic portraits. Finally, Fig.~\ref{fig:vis_imagenet} includes images from the ImageNet-1K dataset, featuring a wide range of objects and scenes with excellent accuracy and visual fidelity. These results emphasize the superior performance of our method in generating high-quality images across different datasets, indicating its potential for broader applications in image synthesis and computer vision tasks.

\begin{figure*}[h]
    \centering
    \includegraphics[width=\textwidth]{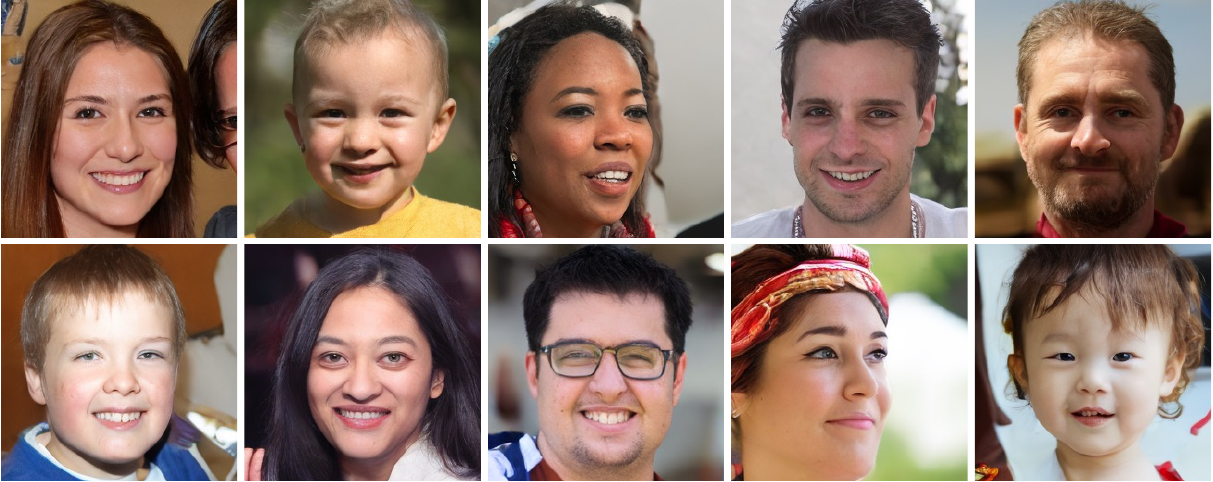}
    \caption{Generated images of FFHQ.}
    \label{fig:vis_ffhq}
\end{figure*}

\end{document}